\documentclass[11pt]{article}
\usepackage[margin=1in]{geometry}
\usepackage[T1]{fontenc}
\usepackage{lmodern}
\usepackage[expansion=false]{microtype}
\usepackage{amsmath,amssymb}
\usepackage{booktabs}
\usepackage{tabularx}
\usepackage{multirow}
\usepackage{graphicx}
\usepackage{xcolor}
\usepackage{tikz}
\usetikzlibrary{positioning,fit,arrows.meta,calc}
\usepackage{pgfplots}
\pgfplotsset{compat=1.17}
\usepackage[round,authoryear]{natbib}
\usepackage{listings}
\definecolor{hobbyblue}{HTML}{2A78D6}
\definecolor{refgray}{HTML}{898781}
\definecolor{inkmuted}{HTML}{52514E}
\usepackage[hidelinks]{hyperref}
\usepackage{caption}
\usepackage{placeins}  % \FloatBarrier before Section 7 and the appendices

\newcolumntype{Y}{>{\raggedright\arraybackslash}X}
\newcommand{\code}[1]{\texttt{#1}}
\title{HobbyLM: A Sparse Mixture-of-Experts Model Family on a Hobby Budget\\[0.4em]
\large Architecture, a 500M Multimodal and Agentic Model, a 1B Adoption, and Image Generation}

\author{
Harish Santhanalakshmi Ganesan\thanks{Designed the HobbyLM architecture and built the 130M and 500M models, all 500M post-training (chat, multimodal, function calling, computer use, diffusion LM), the inference engines and the image model (Sections~\ref{sec:arch}, \ref{sec:500m}, \ref{sec:mm} and~\ref{sec:image}). Correspondence: \texttt{harishsg99@gmail.com}.}
\and
Hariharan M.\,S.\thanks{Adopted the HobbyLM architecture to build HobbyLM-1B: data curation, data pipeline, training infrastructure, pretraining and anneal, evaluation and analysis (Section~\ref{sec:1b}).}
\and
Prabhurajhan R.\footnotemark[2]
}
\date{}

\begin{document}
\maketitle

\begin{abstract}
We describe HobbyLM, a family of small sparse Mixture-of-Experts (MoE) models trained from scratch with a few hundred to a few thousand dollars of rented GPU time.
The core is a decoder-only transformer with fine-grained experts, sigmoid routing with auxiliary-loss-free load balancing, grouped-query attention with per-head QK-normalization, and a Muon/AdamW optimizer split. Each of these choices was selected with a single-variable ablation suite at 130M parameters.
On this architecture we train a 500M-parameter model (169M active) on 40B FineWeb tokens and post-train it into a chat model, a masked-diffusion language model, a function-calling and computer-use agent, and a single ``omni'' checkpoint that reads text, images, video, sound and speech through frozen encoders and small projectors.
We run these models on a laptop CPU with a from-scratch Rust inference engine.
Two of the authors then adopted the architecture without architectural changes to build HobbyLM-1B (1.04B total, 305M active parameters), trained on a 100B-token curated mixture with a high-quality anneal. HobbyLM-1B reaches a 47.5 average on seven zero-shot benchmarks, and its builders also release an instruction-tuned model and an on-device macOS computer-use adapter.
Finally, we describe HobbyLM-Image, a 333M-parameter flow-matching diffusion transformer that generates $1024\times1024$ images in a 32$\times$-compressed latent space, together with a dependency-light CPU implementation of the full text-to-image pipeline.
We report what worked, what did not, and where the small-scale ceilings lie.
We also state for each component which author designed or built it.
\end{abstract}

\begin{center}\small
\begin{tabular}{@{}r l@{}}
\textbf{Models} & \url{https://huggingface.co/rootxhacker}\\
\textbf{Code} & \url{https://github.com/harishsg993010/HobbyLM}\\
\textbf{Demo} & \url{https://huggingface.co/spaces/rootxhacker/HobbyLM-Playground}\\
\textbf{HobbyLM-1B models} & \url{https://huggingface.co/harims95}\\
\textbf{HobbyLM-1B docs} & \url{https://github.com/fuellabs-ai/hobbylm-1b}\\
\end{tabular}
\end{center}

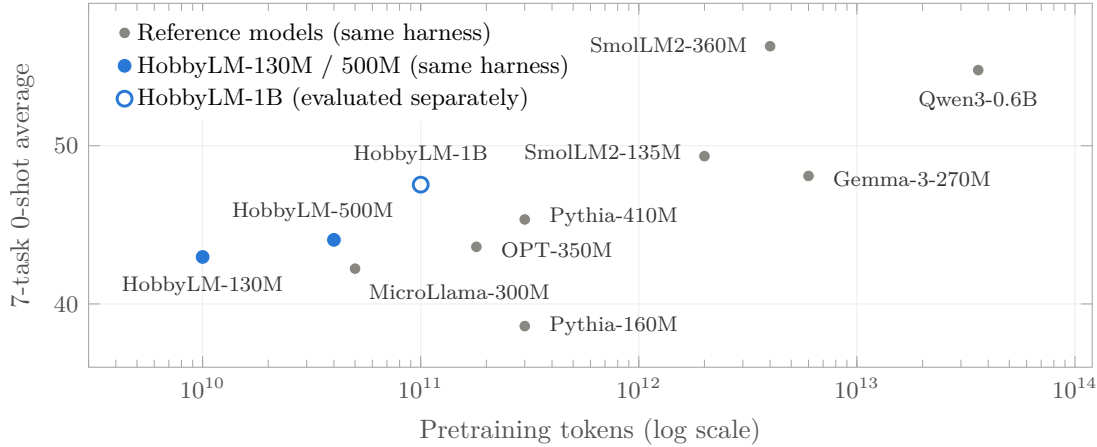
\begin{figure}[t]
\centering
\begin{tikzpicture}
\begin{axis}[
  width=0.9\linewidth, height=6.4cm,
  xmode=log, xmin=3e9, xmax=1.2e14, ymin=36, ymax=59,
  xlabel={Pretraining tokens (log scale)}, ylabel={7-task 0-shot average},
  label style={font=\small, color=inkmuted}, tick label style={font=\footnotesize, color=inkmuted},
  axis line style={black!35}, tick style={black!35},
  grid=major, grid style={draw=black!7},
  legend style={draw=none, fill=white, font=\footnotesize, at={(0.02,0.98)}, anchor=north west},
  legend cell align=left,
  every node near coord/.append style={font=\scriptsize},
]
\addplot[only marks, mark=*, mark size=2.2pt, color=refgray, draw=white, line width=0.4pt] coordinates {
  (4e12,56.29) (3.6e13,54.78) (2e12,49.34) (6e12,48.09) (3e11,45.34) (1.8e11,43.61) (5e10,42.23) (3e11,38.60)};
\addlegendentry{Reference models (same harness)}
\addplot[only marks, mark=*, mark size=2.8pt, color=hobbyblue, draw=white, line width=0.4pt] coordinates {(1e10,42.97) (4e10,44.05)};
\addlegendentry{HobbyLM-130M / 500M (same harness)}
\addplot[only marks, mark=o, mark size=2.8pt, line width=1.1pt, color=hobbyblue] coordinates {(1e11,47.54)};
\addlegendentry{HobbyLM-1B (evaluated separately)}
\begin{scope}[font=\scriptsize, text=black!80]
\node[anchor=north] at (axis cs:1e10,42.4) {HobbyLM-130M};
\node[anchor=south] at (axis cs:3.2e10,44.6) {HobbyLM-500M};
\node[anchor=north west] at (axis cs:5.2e10,41.9) {MicroLlama-300M};
\node[anchor=south] at (axis cs:1e11,48.2) {HobbyLM-1B};
\node[anchor=west] at (axis cs:2.1e11,43.4) {OPT-350M};
\node[anchor=west] at (axis cs:3.5e11,45.5) {Pythia-410M};
\node[anchor=west] at (axis cs:3.5e11,38.6) {Pythia-160M};
\node[anchor=east] at (axis cs:1.75e12,49.6) {SmolLM2-135M};
\node[anchor=east] at (axis cs:3.5e12,56.4) {SmolLM2-360M};
\node[anchor=west] at (axis cs:7e12,47.9) {Gemma-3-270M};
\node[anchor=north] at (axis cs:3.6e13,54.1) {Qwen3-0.6B};
\end{scope}
\end{axis}
\end{tikzpicture}
\caption{Zero-shot average on seven tasks (HellaSwag, OpenBookQA, WinoGrande, ARC-c, ARC-e, BoolQ, PIQA) against pretraining tokens, for base checkpoints of up to about 1B parameters (HobbyLM-1B: 1.04B total, 305M active). Filled points were run through the same evaluation wrapper (Table~\ref{tab:lmeval}). HobbyLM-1B (hollow) was evaluated by its builders on the same tasks with the same metric choices, but with lm-eval 0.4.13 rather than 0.4.9.1 (Appendix~\ref{app:eval}). HobbyLM-130M and -500M outscore MicroLlama-300M, OPT-350M and Pythia-160M, which saw 1.25--30$\times$ more tokens; Pythia-410M (300B tokens) remains ahead of HobbyLM-500M.}\label{fig:teaser}
\end{figure}

% ===========================================================================
\section{Introduction}\label{sec:intro}

Small language models are usually built by distilling or fine-tuning larger released checkpoints. HobbyLM takes the other route. Every model in the family is trained from random initialization, on public data, with rented serverless or on-demand GPUs and no cluster.
We wanted to answer two questions. How far can a sub-billion-parameter sparse MoE get when the architecture and the systems work are chosen carefully? And can one person own every layer of the stack, from the optimizer to the GGUF reader to the desktop application?

Our contributions follow, in the order the work was done. Figure~\ref{fig:teaser} summarizes the base-model results.
\begin{itemize}
\item \textbf{Architecture} (Section~\ref{sec:arch}). A fine-grained sparse MoE decoder whose components were chosen by ablation. QK-normalization turned out to be by far the most important single choice.
\item \textbf{HobbyLM-500M and inference} (Section~\ref{sec:500m}). Pretraining at 130M and 500M; context extension to 8k; chat, masked-diffusion and interpretability variants; and inference through a llama.cpp-compatible export and a from-scratch Rust CPU engine.
\item \textbf{Multimodality and function calling} (Section~\ref{sec:mm}). A TinyLLaVA-style extension to image, video, audio and speech, co-trained into one checkpoint, plus function calling and a text-only computer-use agent driven by accessibility trees.
\item \textbf{HobbyLM-1B} (Section~\ref{sec:1b}). An independent 1B build that adopts the architecture unchanged and adds a curated 100B-token data mixture, stratified shard ordering and an anneal phase. Its builders also release an instruction-tuned model and an on-device macOS computer-use adapter.
\item \textbf{HobbyLM-Image} (Section~\ref{sec:image}). A separate text-to-image diffusion transformer trained in a deep-compression latent space.
\item \textbf{Open artifacts.} Appendix~\ref{app:artifacts} lists all released weights and code for every model, including HobbyLM-1B. Appendices~\ref{app:config}--\ref{app:eval} and~\ref{app:prompts} give training configurations, evaluation settings and prompt formats for the 500M family and the image model; HobbyLM-1B's are in Section~\ref{sec:1b} and Appendices~\ref{app:eval}, \ref{app:sft1b} and~\ref{app:prompts}.
\end{itemize}
Section~\ref{sec:related} places this work among related open models.

\paragraph{Who did what.}
This paper has two distinct bodies of work, and we keep them apart throughout.
\textbf{Harish Santhanalakshmi Ganesan} designed the HobbyLM architecture, ran the ablations, and trained the 130M and 500M models, and also did all 500M post-training, including the multimodal, function-calling and computer-use models, and wrote the inference engines, the desktop application and the image model (Sections~\ref{sec:arch}--\ref{sec:mm} and~\ref{sec:image}).
\textbf{Hariharan M.\,S.} and \textbf{Prabhurajhan R.} took the architecture as specified in Section~\ref{sec:arch} and built HobbyLM-1B in their own code base. They curated the 100B-token training data, built the data pipeline and training infrastructure, ran pretraining and the anneal, and carried out the evaluation and analysis in Section~\ref{sec:1b}.
Table~\ref{tab:provenance} gives the breakdown by component, and Appendix~\ref{app:contrib} gives detailed author contributions.

\begin{table}[t]
\centering\small
\caption{Provenance of each component. ``Adopted'' means the component was taken from the HobbyLM architecture specification and code without modification.}\label{tab:provenance}
\begin{tabularx}{\linewidth}{@{}l Y >{\raggedright\arraybackslash}p{3.1cm}@{}}
\toprule
Component & Content & Designed / built by \\
\midrule
Architecture & MoE decoder, routing, balancing, normalization, optimizer split, presets, ablations & Harish Santhanalakshmi Ganesan \\
130M / 500M pretraining & FineWeb training, context extension, lm-eval wrapper & Harish Santhanalakshmi Ganesan \\
500M post-training & Chat (SFT+DPO), diffusion LM, SAE, multimodal omni model, function calling, computer use & Harish Santhanalakshmi Ganesan \\
Inference & GGUF export, Rust engine \code{hobby-rs}, desktop app \code{hobby-chat}, demo Space & Harish Santhanalakshmi Ganesan \\
HobbyLM-Image & DiT design, data curriculum, training, Rust port & Harish Santhanalakshmi Ganesan \\
\midrule
HobbyLM-1B architecture & The \code{1B} preset of Section~\ref{sec:arch} & Adopted (Harish Santhanalakshmi Ganesan) \\
HobbyLM-1B build & 100B-token mixture, stratified shard order, H200 training, anneal, evaluation, analysis & Hariharan M.\,S., Prabhurajhan R. \\
HobbyLM-1B-Instruct & instruction tuning of HobbyLM-1B; GGUF build; patched CPU runtimes & Hariharan M.\,S., Prabhurajhan R. \\
HobbyLM-1B MLX & MLX implementation for Apple Silicon & Prabhurajhan R. \\
HobbyLM-1B Computer Use v0.1 & Mac computer-use LoRA & Prabhurajhan R. \\
\bottomrule
\end{tabularx}
\end{table}

% ===========================================================================
\section{Background and related work}\label{sec:related}

\paragraph{Small open language models.}
Open model suites below one billion parameters include Pythia~\citep{biderman2023pythia}, OPT~\citep{zhang2022opt} and, more recently, SmolLM2~\citep{allal2025smollm2}, Gemma~3~\citep{gemma2025gemma3} and Qwen3~\citep{yang2025qwen3}.
The recent models train on 2--36T tokens and rely heavily on curated data such as FineWeb-Edu~\citep{penedo2024fineweb}, DCLM~\citep{li2024dclm} and FineMath~\citep{allal2025smollm2}.
MobileLLM~\citep{liu2024mobilellm} argues that deep, thin shapes suit sub-billion models.
HobbyLM works in a far smaller token budget (10--100B tokens) and asks how much careful architecture and systems work can recover. Our training stack borrows from the \code{modded-nanogpt} speedrun~\citep{jordan2024modded}, including the Muon optimizer~\citep{jordan2024muon,liu2025moonlight}.

\paragraph{Sparse Mixture-of-Experts language models.}
Sparsely gated experts~\citep{shazeer2017moe} were scaled up by Switch Transformers~\citep{fedus2022switch} and stabilized by ST-MoE's router z-loss~\citep{zoph2022stmoe}.
DeepSeekMoE~\citep{dai2024deepseekmoe} introduced fine-grained and shared experts, and fine-grained scaling laws~\citep{krajewski2024finegrained} support high granularity.
DeepSeek-V3~\citep{deepseek2024v3} combined sigmoid routing with auxiliary-loss-free balancing~\citep{wang2024auxfree}.
OLMoE~\citep{muennighoff2025olmoe} is the closest open blueprint: a fully open MoE with extensive design-choice ablations at 1B active parameters.
Ling~2.0~\citep{ling2025} uses a routing and normalization layout close to ours, which we exploit for llama.cpp compatibility.
Our contribution is not a new MoE mechanism. It is an ablation-validated combination of these components at 62M--305M active parameters, together with evidence of what transfers to multimodal, agentic and diffusion variants.

\paragraph{Small multimodal and agentic models.}
LLaVA~\citep{liu2023llava,liu2024llava15} established the frozen-encoder-plus-projector recipe, and TinyLLaVA~\citep{zhou2024tinyllava} showed that it works with small language models.
We follow that recipe with SigLIP~\citep{zhai2023siglip,tschannen2025siglip2}, Whisper~\citep{radford2023whisper} and CLAP~\citep{wu2023clap} encoders, and co-train all modalities in one checkpoint.
For function calling we evaluate on BFCL~\citep{patil2025bfcl} and use the open \code{needle} metric code~\citep{cactus_needle}. For GUI agents, much recent work grounds actions in screenshots, for example Aria-UI~\citep{yang2024ariaui}. Our computer-use model instead reads the operating system's accessibility tree as text.

\paragraph{Diffusion language models.}
Masked diffusion language models~\citep{sahoo2024mdlm,nie2025llada} generate by iterative unmasking.
DiffuLLaMA~\citep{gong2025diffullama} adapts autoregressive weights to this objective, and Fast-dLLM~\citep{wu2025fastdllm} accelerates decoding with caching and parallel commitment.
We apply the adaptation recipe at 500M and measure where parallel decoding pays off.

\paragraph{Efficient text-to-image generation.}
Diffusion transformers~\citep{peebles2023dit} trained with rectified-flow objectives~\citep{liu2023rectflow,lipman2023flowmatching,esser2024sd3} are now the standard text-to-image backbone.
SANA~\citep{xie2025sana} and its deep-compression autoencoder DC-AE~\citep{chen2025dcae} cut the token count by compressing $32\times$ per spatial dimension.
HobbyLM-Image builds on this latent space to train a 1024px generator on a hobby budget.

% ===========================================================================
\section{Architecture}\label{sec:arch}
\emph{Designed by Harish Santhanalakshmi Ganesan}

\begin{figure}[t]
\centering
\begin{tikzpicture}[
  font=\small,
  box/.style={draw, rounded corners=2pt, minimum height=0.62cm, inner sep=3pt, align=center},
  arr/.style={-{Latex[length=2mm]}, thick},
  node distance=0.32cm]
% left: block stack
\node[box, minimum width=4.4cm] (emb) at (0,0) {Token embedding (tied)};
\node[box, minimum width=4.4cm] (l0) at (0,1.1) {Layer 0: attention + dense SwiGLU};
\node[box, minimum width=4.4cm] (l1) at (0,2.2) {Layers $1..L{-}1$: attention + MoE};
\node[box, minimum width=4.4cm] (fn) at (0,3.3) {Final RMSNorm $\rightarrow$ tied LM head};
\draw[arr] (emb) -- (l0); \draw[arr] (l0) -- (l1); \draw[arr] (l1) -- (fn);
% right top: attention
\node[box, minimum width=5.6cm] (qkv) at (8.2,4.6) {fused $W_{qkv}$: $Q$ ($n_q$ heads), $K,V$ ($n_{kv}$ heads)};
\node[box, minimum width=5.6cm] (qkn) at (8.2,3.7) {per-head RMSNorm on $Q$ and $K$ (QK-norm)};
\node[box, minimum width=5.6cm] (rope) at (8.2,2.8) {RoPE $\rightarrow$ causal SDPA (GQA)};
\draw[arr] (qkv) -- (qkn); \draw[arr] (qkn) -- (rope);
\node[draw, dashed, rounded corners, fit=(qkv)(qkn)(rope), inner sep=5pt, label={[font=\small\bfseries]above:Attention}] (att) {};
% right bottom: MoE
\node[box, minimum width=5.6cm] (rt) at (8.2,1.2) {router (fp32): $s=\sigma(xW_g)$};
\node[box, minimum width=5.6cm] (sel) at (8.2,0.3) {select top-$k$ of $s+b$; gate with raw $s$};
\node[box] (ex) at (6.75,-0.75) {$k$ of $E$ routed experts};
\node[box] (sh) at (9.85,-0.75) {shared expert};
\draw[arr] (rt) -- (sel); \draw[arr] (sel.south -| ex.north) -- (ex.north);
\node[draw, dashed, rounded corners, fit=(rt)(sel)(ex)(sh), inner sep=5pt, label={[font=\small\bfseries]below:Sparse MoE FFN}] (moe) {};
\draw[dashed] (l1.east) -- (att.west); \draw[dashed] (l1.east) -- (moe.west);
\end{tikzpicture}
\caption{The HobbyLM decoder. Every layer is pre-norm. Layer~0 uses a dense SwiGLU FFN; all later layers use a sparse MoE with $E$ routed SwiGLU experts, top-$k$ sigmoid routing, an auxiliary-loss-free selection bias $b$, and (in the 500M and 1B presets) one shared expert.}\label{fig:arch}
\end{figure}
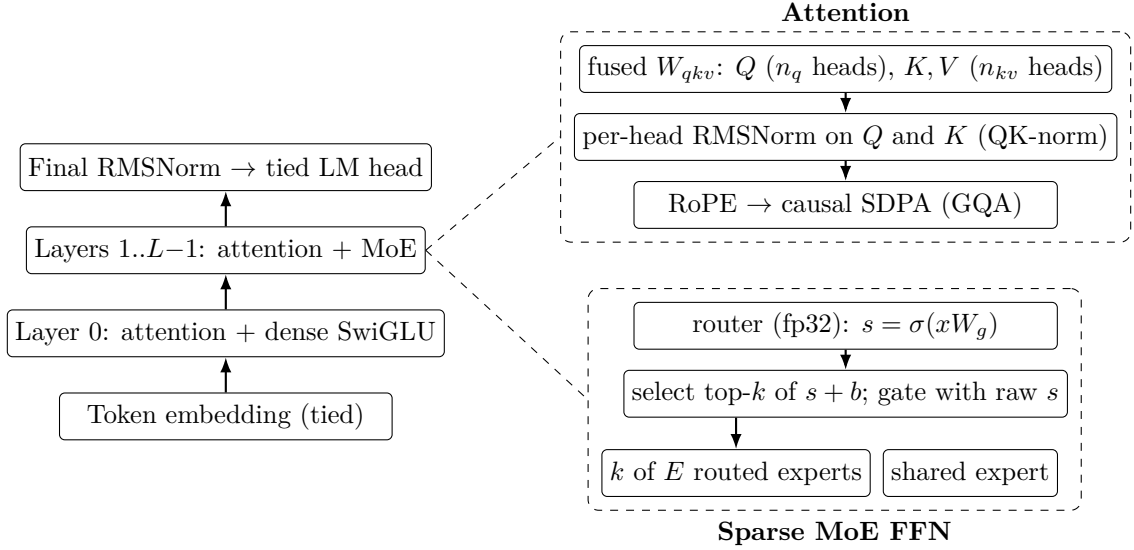

HobbyLM is a decoder-only transformer~\citep{vaswani2017attention} with pre-normalization RMSNorm~\citep{zhang2019rmsnorm}, rotary position embeddings~\citep{su2024roformer}, tied input/output embeddings~\citep{press2017tying} and the GPT-2 byte-level BPE tokenizer~\citep{radford2019gpt2}. The vocabulary is 50,257 tokens padded to 50,304. The unused padded slots later serve as modality sentinels (Section~\ref{sec:mm}).
The feed-forward network of every layer except the first is a fine-grained sparse MoE~\citep{shazeer2017moe,fedus2022switch,dai2024deepseekmoe}. Figure~\ref{fig:arch} gives an overview.
We drew the design from the recent small-MoE literature (DeepSeekMoE/V3~\citep{dai2024deepseekmoe,deepseek2024v3}, OLMoE~\citep{muennighoff2025olmoe}, Qwen3~\citep{yang2025qwen3}) and from fine-grained MoE scaling laws~\citep{krajewski2024finegrained}. We then kept or dropped each disputed choice according to the ablations in Section~\ref{sec:ablations}.

\subsection{Attention}
Attention uses grouped-query attention (GQA)~\citep{ainslie2023gqa} with a single fused $QKV$ projection and no biases.
The head dimension is decoupled from the model width, following Qwen3~\citep{yang2025qwen3}: the 500M model uses 12 query heads of size 128 ($12\times128=1536$) on a 768-wide residual stream.
We apply RMSNorm per head to $Q$ and $K$ \emph{before} RoPE (QK-norm~\citep{henry2020qknorm,dehghani2023vit22b}). This bounds the attention logits.
In our ablations removing QK-norm was by far the most damaging single change (Table~\ref{tab:ablations}). The same intervention later fixed the training divergence of the image model (Section~\ref{sec:image}).
Keys and values are expanded to the query heads and passed to scaled-dot-product attention with a causal mask. The masked-diffusion variant (Section~\ref{sec:diffusion}) is the only exception and uses no mask.

\subsection{Sparse MoE feed-forward}
Each expert is a SwiGLU network~\citep{shazeer2020glu} with a narrow hidden width $f$, so granularity $G=f_{\text{dense}}/f$ is high~\citep{krajewski2024finegrained}.
Following DeepSeekMoE~\citep{dai2024deepseekmoe}, the first layer stays dense, because early-layer routing is unstable.
For a token representation $x\in\mathbb{R}^d$ the router computes sigmoid scores in fp32,
\begin{equation}
s_i = \sigma\!\left(x^\top w_i\right), \qquad i=1,\dots,E,
\end{equation}
and selects experts using biased scores, as in DeepSeek-V3's auxiliary-loss-free balancing~\citep{wang2024auxfree,deepseek2024v3}:
\begin{equation}
\mathcal{T}(x) = \operatorname{TopK}_i\big(s_i + b_i,\;k\big), \qquad
y = \sum_{i\in\mathcal{T}(x)} s_i\,\mathrm{FFN}_i(x) \;+\; \mathrm{FFN}_{\text{shared}}(x).
\end{equation}
The bias $b$ only affects \emph{which} experts are selected. The gate weights are the raw scores $s_i$, so the balancing signal does not compete with the language-modeling gradient.
We do not renormalize the selected gates to sum to one. Renormalization was the default in our initial design, but the ablation (Table~\ref{tab:ablations}) favored leaving it off. OLMoE likewise weights the selected experts by their unrenormalized routing probabilities~\citep{muennighoff2025olmoe}.
After each forward pass in training, the bias moves by a fixed step toward balanced load,
\begin{equation}
b_i \leftarrow b_i + u\cdot\operatorname{sign}\!\big(\bar{c} - c_i\big),
\end{equation}
where $c_i$ is the number of tokens routed to expert $i$, $\bar c$ is the balanced count and $u=10^{-3}$. Under data parallelism the bias is averaged across ranks after every optimizer step. Bias updates are frozen for the final 5\% of training.
Because the update runs on every forward pass, the effective step per optimizer step is $u$ times the number of gradient-accumulation micro-batches.
We keep two small regularizers. The first is a Switch-style balance term $\alpha E\sum_i f_i P_i$ with $\alpha=10^{-3}$~\citep{fedus2022switch}. The second is the ST-MoE router z-loss~\citep{zoph2022stmoe}, also with coefficient $10^{-3}$. We add a z-loss of $10^{-4}$ on the final logits~\citep{chowdhery2023palm}.

\paragraph{Dropless expert computation.}
Routing has no capacity limit and drops no tokens.
The (token, expert) pairs are sorted by expert, and the gathered rows go through a single grouped GEMM per projection (\code{torch.\_grouped\_mm}) against the experts' weights, stored as 3-D tensors $W_{13}\in\mathbb{R}^{E\times d\times 2f}$ and $W_2\in\mathbb{R}^{E\times f\times d}$. The outputs are scaled by their gates and scattered back with an indexed add.
The shared expert runs on every token.
At this scale no expert parallelism is needed: the 1B model fits on a single GPU.

\paragraph{Initialization.}
All weights are initialized from $\mathcal{N}(0, 0.02^2)$. Residual output projections (attention output and every expert's down-projection) are additionally scaled by $1/\sqrt{2L}$. The router is scaled by a further $0.1$, following the reduced initialization scale that Switch Transformers recommends for stability~\citep{fedus2022switch} and ST-MoE adopts~\citep{zoph2022stmoe}.

\subsection{Optimizer}
We use Muon~\citep{jordan2024muon,liu2025moonlight} for every matrix-shaped hidden parameter, including the 3-D expert stacks. Each expert's matrix is orthogonalized independently with a batched Newton--Schulz iteration (5 steps, bf16).
The update is rescaled by $0.2\sqrt{\max(m,n)}$ to match AdamW's update RMS, as in Moonlight~\citep{liu2025moonlight}, and uses Nesterov momentum $0.95$ and decoupled weight decay $0.1$.
The router, the tied embedding/output matrix and all vector parameters use AdamW~\citep{loshchilov2019adamw} with $\beta=(0.9,0.95)$.
We keep the router on AdamW because orthogonalization flattens the singular values of the gate matrix and so would discard routing signal.
Learning rates are $0.02$ (Muon) and $3\times10^{-4}$ (AdamW) at a 262,144-token batch, scaled by $\sqrt{B/262{,}144}$ for larger batches.
The schedule is warmup--stable--decay~\citep{hu2024minicpm}: 100 warmup steps, a constant phase, then linear decay over the last 40\% of training to $0.1\times$ the peak.
Training uses fp32 master weights, bf16 autocast, TF32 matmuls and an fp32 router.
The training stack borrows much of its engineering from \code{modded-nanogpt}~\citep{jordan2024modded}, including the data shard format and loader, on-device loss accumulation and asynchronous host-to-device prefetch.

\subsection{Model presets}
Table~\ref{tab:presets} lists the three presets. Parameter counts come from instantiating each configuration.
The models are deep and thin, as MobileLLM recommends for sub-billion models~\citep{liu2024mobilellm}, and their total-to-active parameter ratios are 2.3--3.4$\times$.

\begin{table}[t]
\centering\small
\caption{HobbyLM presets. $G$ is the dense-to-expert FFN width ratio. All presets use 1 dense leading layer, RMSNorm with $\epsilon=10^{-6}$, RoPE $\theta=10^4$ (raised to $10^6$ for the 8k-context 500M), tied embeddings and vocabulary 50,304.}\label{tab:presets}
\begin{tabular}{@{}l r r c r r r r r r@{}}
\toprule
Preset & $d$ & $L$ & $n_q/n_{kv}\times d_h$ & dense FFN & $E\times f$ & top-$k$ & shared & Total & Active \\
\midrule
130M & 512 & 12 & $8/2\times64$ & 1536 & $32\times192$ & 8 & 0 & 140.0M & 62.1M \\
500M & 768 & 16 & $12/3\times128$ & 2304 & $36\times320$ & 6 & 1 & 500.8M & 169.0M \\
1B & 1024 & 20 & $16/8\times128$ & 2816 & $64\times224$ & 8 & 1 & 1037.1M & 305.0M \\
\bottomrule
\end{tabular}
\end{table}

\subsection{Ablations}\label{sec:ablations}
We tested each contested choice with a single-knob ablation at 130M (32 experts, top-4 baseline at the time) for 3,000 steps on FineWeb. All runs were launched in parallel, one per H100. Table~\ref{tab:ablations} reports the validation losses.
QK-norm dominates: removing it raises the loss by 0.80 nats.
The router z-loss and auxiliary-loss-free balancing each matter at the 0.04 level, and so does keeping at least 32 experts.
Sigmoid and softmax gating were indistinguishable, and so were 0 and 1 shared experts at 32 experts.
We adopted the two winning changes into the defaults: top-8 routing for the 130M preset and no top-$k$ renormalization.
We kept a shared expert in the larger presets, which have more experts, following the literature~\citep{dai2024deepseekmoe}.

\begin{table}[t]
\centering\small
\caption{Single-knob ablations at 130M, 3,000 steps, FineWeb validation loss (baseline 3.6688).}\label{tab:ablations}
\begin{tabular}{@{}l l r r@{}}
\toprule
Variant & Change from baseline & Val loss & $\Delta$ \\
\midrule
\code{topk8} & top-8 instead of top-4 & 3.6441 & $-0.025$ \\
\code{no\_renorm} & no top-$k$ gate renormalization & 3.6542 & $-0.015$ \\
\code{softmax} & softmax instead of sigmoid gating & 3.6677 & $-0.001$ \\
\code{shared1} & add one shared expert & 3.6678 & $-0.001$ \\
\code{baseline} & -- & 3.6688 & 0 \\
\code{scale\_emb} & multiply embeddings by $\sqrt{d}$ & 3.6775 & $+0.009$ \\
\code{aux\_loss} & classic auxiliary loss instead of aux-free bias & 3.7048 & $+0.036$ \\
\code{no\_zloss} & remove router z-loss & 3.7097 & $+0.041$ \\
\code{experts16} & 16 experts instead of 32 & 3.7166 & $+0.048$ \\
\code{no\_qknorm} & remove QK-norm & 4.4715 & $+0.803$ \\
\bottomrule
\end{tabular}
\end{table}

\subsection{Throughput options}\label{sec:fusedce}
We ported several throughput techniques from \code{modded-nanogpt} behind flags and evaluated them for speed and quality (Table~\ref{tab:speed}).
The most useful was a chunked cross-entropy (\code{fused\_ce}). It computes the vocabulary projection and the loss in row chunks under activation checkpointing, so the full $(T\times V)$ fp32 logit tensor is never materialized. The loss and gradients are bit-identical to the baseline on CPU, and peak memory drops by 21\%. This allows roughly twice the micro-batch size: at micro-batch 32 on a 1B probe, the baseline ran out of memory, while \code{fused\_ce} reached about 74k tokens/s on one H100.
The Polar Express orthogonalization schedule~\citep{amsel2025polar} was neutral at our horizon.
An FP8 output head gave no speedup and, in our implementation, passed no gradient through its backward pass. We dropped it.

\begin{table}[t]
\centering\small
\caption{Throughput options on a 1B speed probe (one H100, micro-batch 8, sequence 1024) and quality on a 130M, 1,000-step run.}\label{tab:speed}
\begin{tabular}{@{}l r r r r@{}}
\toprule
Option & ms/step & tokens/s & Peak GB & 130M val loss \\
\midrule
baseline & 222.5 & 36,821 & 33.3 & 4.0739 \\
\code{fused\_ce} & 209.2 & 39,166 & 26.4 & 4.0681 \\
\code{polar} & 224.8 & 36,445 & 33.3 & 4.0790 \\
\code{fused\_ce}+\code{polar} & 210.1 & 38,988 & 26.4 & 4.0855 \\
FP8 head & 223.8 & 36,605 & 33.8 & 10.90 (no gradient) \\
\bottomrule
\end{tabular}
\end{table}

% ===========================================================================
\section{HobbyLM-500M: pretraining, variants and inference}\label{sec:500m}
\emph{Work by Harish Santhanalakshmi Ganesan}

\subsection{Pretraining}
Both models were trained on FineWeb~\citep{penedo2024fineweb}, pre-tokenized with GPT-2 into \code{modded-nanogpt} binary shards, on 8$\times$H100 with PyTorch DDP~\citep{li2020ddp} and \code{torch.compile}~\citep{ansel2024pytorch2}. Each optimizer step covers $2^{20}$ tokens (1,048,576).
\begin{itemize}
\item \textbf{130M:} 9,537 steps (${\approx}10$B tokens), about 742\,ms/step, about 2 hours. Final validation loss 3.3016.
\item \textbf{500M:} 38,147 steps (${\approx}40$B unique tokens from FineWeb-100B), about 1.21\,s/step, about 13 hours. Final validation loss \textbf{3.0281}.
\end{itemize}
Peak learning rates were Muon $0.04$ and AdamW $6\times10^{-4}$ (the $\sqrt{4}=2$ batch scaling).

\paragraph{Context extension.}
For multimodal inputs the 500M model needed a longer context. We continued pretraining at sequence length 2048 for 2,000 steps (${\approx}2.1$B tokens, learning rate $\times0.5$, $\theta=10^4$). Validation loss \emph{improved} to 2.999 at 2048.
A second extension to 8,192 tokens raised RoPE $\theta$ to $10^6$ and reached validation loss 2.94 at 8k.

\subsection{Zero-shot evaluation}
We evaluate with the EleutherAI LM Evaluation Harness~\citep{gao2024lmeval,biderman2024lessons} (v0.4.9.1) through a wrapper that scores log-likelihoods with the MoE and the GPT-2 tokenizer. The wrapper encodes context and continuation jointly, as the harness does for Hugging Face models.
We report \code{acc\_norm} where defined and \code{acc} otherwise, all 0-shot, on HellaSwag~\citep{zellers2019hellaswag}, OpenBookQA~\citep{mihaylov2018obqa}, WinoGrande~\citep{sakaguchi2020winogrande}, ARC~\citep{clark2018arc}, BoolQ~\citep{clark2019boolq} and PIQA~\citep{bisk2020piqa}.
We ran all reference models through the \emph{same} wrapper and task set. We checked the wrapper by reproducing published averages for two models: MicroLlama~\citep{microllama} 42.23 against 42.36 on its model card, and TinyLlama-1.1B-3T~\citep{zhang2024tinyllama} 52.75 against 52.99.
Table~\ref{tab:lmeval} shows that the ranking follows pretraining tokens more closely than parameters.
The models ahead of HobbyLM-500M were trained on 300B to 36T tokens, 7.5--900$\times$ our budget.
Among models trained on at most 300B tokens, the 130M model (10B tokens) outscores MicroLlama-300M (50B), OPT-350M (180B) and Pythia-160M (300B).
On two further tasks, the 500M model scores 40.0 on LAMBADA~\citep{paperno2016lambada} and 70.3 on SciQ~\citep{welbl2017sciq}.

\begin{table}[t]
\centering\small
\caption{0-shot results, all base (not instruction-tuned) checkpoints, all run through the same harness wrapper. Reference models: SmolLM2~\citep{allal2025smollm2}, Qwen3~\citep{yang2025qwen3}, Gemma~3~\citep{gemma2025gemma3}, Pythia~\citep{biderman2023pythia}, OPT~\citep{zhang2022opt}, GPT-2~\citep{radford2019gpt2} and MicroLlama~\citep{microllama}.}\label{tab:lmeval}
\setlength{\tabcolsep}{4pt}
\begin{tabular}{@{}l r r r r r r r r r r@{}}
\toprule
Model & Params & Tokens & Hella & OBQA & Wino & ARC-c & ARC-e & BoolQ & PIQA & Avg \\
\midrule
SmolLM2-360M & 360M & 4T & 56.41 & 38.00 & 59.35 & 38.48 & 68.22 & 62.08 & 71.49 & 56.29 \\
Qwen3-0.6B & 600M & 36T & 53.81 & 34.60 & 58.56 & 38.31 & 58.08 & 69.82 & 70.29 & 54.78 \\
SmolLM2-135M & 135M & 2T & 43.13 & 33.20 & 52.25 & 29.44 & 58.59 & 60.37 & 68.39 & 49.34 \\
Gemma-3-270M & 270M & 6T & 41.28 & 30.60 & 53.43 & 28.33 & 57.07 & 57.77 & 68.12 & 48.09 \\
Pythia-410M & 410M & 300B & 40.14 & 28.80 & 52.96 & 23.98 & 45.45 & 58.50 & 67.52 & 45.34 \\
\textbf{HobbyLM-500M} & 501M/169M & 40B & 41.56 & 29.80 & 51.30 & 22.44 & 42.76 & 50.98 & 69.53 & \textbf{44.05} \\
OPT-350M & 350M & 180B & 36.55 & 28.40 & 52.80 & 24.74 & 40.49 & 57.61 & 64.69 & 43.61 \\
\textbf{HobbyLM-130M} & 140M/62M & 10B & 32.54 & 28.20 & 52.17 & 23.46 & 37.71 & 61.31 & 65.40 & \textbf{42.97} \\
MicroLlama-300M & 300M & 50B & 34.22 & 30.40 & 51.30 & 23.21 & 39.23 & 52.57 & 64.69 & 42.23 \\
GPT-2 (124M) & 124M & -- & 31.15 & 27.40 & 51.46 & 22.53 & 39.60 & 49.94 & 62.24 & 40.62 \\
Pythia-160M & 160M & 300B & 30.21 & 26.60 & 49.64 & 24.83 & 36.95 & 42.42 & 59.52 & 38.60 \\
\bottomrule
\end{tabular}
\end{table}

\subsection{Chat model}
We fine-tuned the 500M model on about 1.5M SmolTalk conversations~\citep{allal2025smollm2} (\code{smol-smoltalk} and the conversational, non-reasoning subsets of SmolTalk2), with the loss on assistant turns only. The run lasted 28k steps, ending at loss 1.50.
We then applied DPO~\citep{rafailov2023dpo} on UltraFeedback preference pairs~\citep{cui2024ultrafeedback} with $\beta=0.1$ and learning rate $5\times10^{-7}$, which raised preference accuracy from 0.50 to 0.64. This follows the SmolLM2 recipe.
The chat model averages 42.5 on the seven tasks, against 44.0 for the base model. ARC-challenge and WinoGrande held or improved; BoolQ dropped the most (51.0 to 44.4).
At decoding time a repetition penalty of 1.3 removes most of the looping we observed with greedy decoding.

\subsection{Masked-diffusion variant}\label{sec:diffusion}
We converted the autoregressive 500M model into a masked-diffusion language model in the style of LLaDA and MDLM~\citep{nie2025llada,sahoo2024mdlm}. As in DiffuLLaMA~\citep{gong2025diffullama}, we started from the autoregressive weights and replaced the causal mask with full bidirectional attention.
Training draws $t\sim\mathcal U(\epsilon,1)$ per sequence, masks each token independently with probability $t$, and minimizes the masked-token cross-entropy weighted by $1/t$. A free vocabulary slot serves as the mask token.
After about 20B tokens (8$\times$B200) the validation loss was 3.52.
A chat variant was then fine-tuned on SmolTalk by masking only the assistant response.
We found decoding settings matter more than we expected: greedy commitment with more steps, a repetition penalty of 1.4--1.5 and low-confidence remasking turned incoherent samples into fluent multi-sentence prose.
We implemented the semi-autoregressive block sampler with a cached prefix in our Rust engine (Section~\ref{sec:infer}).
On an H100 the diffusion model generates 117.7 tokens/s (128 tokens, 32 steps), about 2.7$\times$ the autoregressive model's ${\sim}44$ tokens/s. For comparison, Fast-dLLM reports 2--3.6$\times$ from KV caching alone, though against a vanilla diffusion baseline rather than an autoregressive one~\citep{wu2025fastdllm}.
On a compute-bound laptop CPU the order reverses (${\sim}6.5$ against ${\sim}52$ tokens/s). Whether diffusion decoding wins depends on whether the hardware is memory-bound or compute-bound.
The diffusion chat model adopts the conversational register but hallucinates more than the autoregressive chat model.

\subsection{Interpretability with sparse autoencoders}
We trained a top-$k$ sparse autoencoder~\citep{gao2025sae} with 12,288 latents and $k=32$ on the residual stream after layer~8 of the 8k-context base model. Activations were streamed online from FineWeb (50M tokens), with no activation cache.
The SAE explains 97.5--98.4\% of the variance at $L_0=32$, with about 20 dead latents.
Inspection of top-activating tokens shows concept features (e.g.\ music, data, devices), synonym clusters (\emph{ignored / overlooked / neglected}) and syntactic patterns (``\% off'', ``Table $N$'').
We labeled the features automatically from their top tokens. They are shown together with live expert-routing maps in our public demo.

\subsection{Inference}\label{sec:infer}

\paragraph{llama.cpp compatibility.}
The architecture has the same structure as llama.cpp's~\citep{gerganov2023llamacpp} \code{bailingmoe2} model (Ling 2.0~\citep{ling2025}): GQA with per-head QK-norm, sigmoid-gated experts with a selection bias, a shared expert and leading dense layers.
We export to GGUF~\citep{gguf_spec} with that tensor layout. The 3-D expert stacks are transposed to llama.cpp's $\{d,f,E\}$ order and the selection bias is stored as \code{exp\_probs\_b}.
A one-line patch adds support for a head dimension decoupled from the model width.
Generations matched the PyTorch reference token for token.
Quantizing to Q4\_K\_M reduced the file from 2.16\,GB to 362\,MB and raised CPU decoding from about 13 to 43 tokens/s.
The vision and speech encoders export to llama.cpp's multimodal projector formats (\code{phi4} for SigLIP2 and \code{voxtral} for Whisper).
Release GGUFs declare a custom \code{hobbylm} architecture name; the \code{bailingmoe2} export remains available.

\paragraph{\code{hobby-rs}, a from-scratch CPU engine.}
\code{hobby-rs} is a Rust inference engine with no llama.cpp, ONNX or Python dependency at run time. Its parts:
\begin{itemize}
\item a GGUF v3 reader that takes every hyperparameter from the file's metadata;
\item a GPT-2 BPE tokenizer built from the embedded vocabulary;
\item a KV cache;
\item sparse MoE execution that computes only the selected experts, batching prefill tokens per expert;
\item a SIMD dot product and an int8$\times$int8 Q8\_0 kernel in the style of ggml's \code{maddubs}.
\end{itemize}
Its output is token-for-token identical to llama.cpp.
On a 15\,W Intel i7-1250U laptop, enabling native AVX2 code generation and the int8 kernel raised Q8 decoding from about 28 to about 52 tokens/s.
The engine also implements the diffusion samplers (fixed-schedule and confidence-threshold parallel decoding) and accepts precomputed image or speech embeddings. These let it run the multimodal model on a CPU.

\paragraph{\code{hobby-chat} and the demo.}
\code{hobby-chat} is a desktop application built with Tauri~\citep{tauri} that embeds \code{hobby-rs}.
It runs the SigLIP2 and Whisper encoders on the CPU through \code{candle}~\citep{candle}, and serves an OpenAI-compatible HTTP endpoint that our agent loops use.
It also drives a Windows UI-Automation~\citep{microsoft_uia} computer-use loop (Section~\ref{sec:cua}) and connects to tools through an MCP client.
A public Gradio Space provides chat with every variant, image QA, image generation, per-token MoE routing visualizations and SAE feature browsing.

% ===========================================================================
\section{Multimodality and function calling}\label{sec:mm}
\emph{Work by Harish Santhanalakshmi Ganesan}

\subsection{Design}
We extend the 500M model in the style of LLaVA and TinyLLaVA~\citep{liu2023llava,zhou2024tinyllava}. Frozen encoders produce feature sequences, a small trainable two-layer GELU MLP~\citep{hendrycks2016gelu} projects each sequence to the model width, and the projected features replace a sentinel token in the input embeddings.
The sentinels live in the unused padded vocabulary slots (IDs 50257--50262), so we never changed the tokenizer.
Table~\ref{tab:encoders} lists the encoders.
\begin{itemize}
\item Speech features are stacked two frames at a time, halving Whisper's 1,500 frames to 750 tokens.
\item Video reuses the image path on sampled frames, keeping the raw per-frame tokens. Pooled frame tokens fell outside the projector's training distribution and produced unusable output.
\end{itemize}
The encoder features are pre-computed and cached wherever possible, so training never runs the vision transformer.
During splicing, every feature position has its training target masked except the last. The last feature position predicts the token that follows the sentinel.

\begin{table}[t]
\centering\small
\caption{Frozen encoders and projectors of the omni model: SigLIP2~\citep{tschannen2025siglip2,zhai2023siglip}, CLAP~\citep{wu2023clap} with an HTSAT backbone~\citep{chen2022htsat}, and Whisper~\citep{radford2023whisper}.}\label{tab:encoders}
\begin{tabularx}{\linewidth}{@{}l l Y r@{}}
\toprule
Modality & Frozen encoder & Tokens per input & Projector \\
\midrule
Image (early) & SigLIP2 so400m/14 @384 & 729 $\times$ 1152 & 1.48M \\
Image (from joint11) & SigLIP2 so400m/16 @512 & 1024 $\times$ 1152 & 1.48M \\
Video & same as image, per frame & 729 or 1024 per frame & shared \\
Sound & CLAP HTSAT & flattened HTSAT feature map, $\times$768 & 1.18M \\
Speech & Whisper-small encoder & 750 $\times$ 1536 (2-frame stacking) & 1.77M \\
\bottomrule
\end{tabularx}
\end{table}

\subsection{Staged training}
All stages ran on Modal, almost all on 8$\times$H100.
\begin{enumerate}
\item \textbf{Vision alignment.} Projector only, on the LLaVA-Pretrain LCS-558K captions~\citep{liu2023llava,schuhmann_laion_cc_sbu}, 1,500 steps. The loss fell from 6.66 to 3.02.
\item \textbf{Vision instruction tuning.} We unfroze the model and fine-tuned on LLaVA-Instruct-150K~\citep{liu2023llava} with COCO images~\citep{lin2014coco}, using AdamW at $2\times10^{-5}$ (model) and $10^{-4}$ (projector) with a cosine schedule, for one epoch (2,500 steps, final loss 1.58).
\item \textbf{Sound.} Audio-projector-only alignment on Clotho~\citep{drossos2020clotho} (3,839 clips, 1,200 steps; loss 4.46 to 2.77). The model distinguishes, for example, birds, crickets, engines and pouring water.
\item \textbf{Speech.} Speech-projector-only ASR alignment on LibriSpeech train-clean-100~\citep{panayotov2015librispeech} (about 12K utterances, 1,500 steps). This was followed by spoken question answering on VoiceAssistant-400K~\citep{xie2024miniomni}, with the model unfrozen.
\item \textbf{Joint co-training.} A single interleaved cycle over all modality ``paths'', so no modality drifts while others are trained. We grew the cycle across eight runs (Table~\ref{tab:joint}). Each run resumes from the previous one.
\end{enumerate}

\begin{table}[t]
\centering\small
\caption{Joint co-training lineage of the omni model. Each run adds paths to the cycle and resumes from its predecessor.}\label{tab:joint}
\begin{tabularx}{\linewidth}{@{}l Y@{}}
\toprule
Run & Added \\
\midrule
joint5 & image, video, sound, speech (spoken QA), text \\
joint6 & single-shot tool calls \\
joint7 & text agent loop (query $\to$ call $\to$ result $\to$ answer) and spoken agent loop (TTS'd queries) \\
joint8 & text must-call up-weighted \\
joint9 & Nemotron agentic SFT (v2) and Nemotron VLM captioning~\citep{nvidia_nemotron_vlm}, streamed \\
joint10 & OCR (rendered text $\to$ markdown) \\
joint11 & 8k context ($\theta=10^6$) and SigLIP2 @512 (1,024 tokens per image) \\
joint12 & SmolTalk chat, mobile actions~\citep{google_mobile_actions}, Aria-UI desktop grounding~\citep{yang2024ariaui}; 18 paths, 13 modalities/tasks \\
\bottomrule
\end{tabularx}
\end{table}

\paragraph{Results.}
Table~\ref{tab:vlm} lists the multimodal results.
POPE~\citep{li2023pope}, GQA~\citep{hudson2019gqa} and VQAv2~\citep{goyal2017vqav2} were run with lmms-eval~\citep{zhang2024lmmseval} through a custom model adapter.
Our chat-trained model answers in sentences, so strict single-word exact match under-scores it. We report a normalized exact match (leading yes/no for POPE; the salient content word not already in the question for GQA and VQAv2) and, for the final model, also containment scoring.
POPE exposes the clearest weakness: the stage-2 model answers ``yes'' to every probe, so accuracy is 50\% and recall is 1.0.
Speech recognition with only the 1.77M-parameter projector trained, and the language model frozen, reaches 28.54\% WER on LibriSpeech test-clean. Most errors are homophones (\emph{francs}$\to$\emph{franks}).
Two zero-shot behaviors appeared without any training on the combination:
\begin{itemize}
\item The model grounds a spoken question in an image (``how many people?'' $\to$ ``three girls'').
\item Video works through the image path with no video training data.
\end{itemize}

\begin{table}[t]
\centering\small
\caption{Multimodal results of the 500M model. For the stage-2 model, GQA and VQAv2 use 2K-question subsets and POPE the full 9K set.}\label{tab:vlm}
\begin{tabular}{@{}l l r@{}}
\toprule
Checkpoint & Benchmark (scoring) & Score \\
\midrule
\multirow{3}{*}{Vision stage 2 (384px, 2k ctx)} & POPE accuracy / F1 & 50.0 / 66.7 \\
 & GQA (normalized exact match) & 27.85 \\
 & VQAv2 val (normalized exact match) & 37.29 \\
\midrule
\multirow{2}{*}{joint12 (512px, 8k ctx)} & GQA (strict / containment) & 20.6 / 39.2 \\
 & VQAv2 val (strict / containment) & 28.2 / 47.0 \\
\midrule
Speech stage 1 (projector only) & LibriSpeech test-clean WER $\downarrow$ & 28.54 \\
\bottomrule
\end{tabular}
\end{table}

\subsection{Function calling}\label{sec:tools}
We train single-shot function calling in the prompt format \code{TOOLS: [schemas] \textbackslash n USER: \ldots\ ASSISTANT:} followed by a JSON list of calls. We score it with the metrics of the open \code{needle} benchmark code~\citep{cactus_needle}: JSON parse rate, name F1, argument and value accuracy, parameter hallucination and exact match.
The first model was fine-tuned on 268K single-call examples from Nemotron-Agentic-v1~\citep{nvidia_nemotron_agentic} for 4,000 steps. These were the rows whose first tool call fits in 2,048 tokens.
We then tried two levers (Table~\ref{tab:tools}).
\begin{itemize}
\item \textbf{Grammar-constrained decoding.} A character-level JSON state machine restricts tokens to valid tool names and argument keys and leaves values free. It removed parameter hallucination (3.8\% to 0.0\%) but left exact match unchanged. The remaining errors were semantic: a valid but wrong tool, or a value formatted differently from the reference (\code{us} vs.\ \code{US}).
\item \textbf{Token-weighted loss.} Weighting the loss 3$\times$ on names, 2$\times$ on values and 1.5$\times$ on keys gave about one point everywhere.
\end{itemize}
Adding cleaner data mattered more. Training on Nemotron, BitAgent~\citep{bitagent_toolcalling} and a small multi-call set~\citep{interstellar_toolcalls} (about 816K examples, 6,000 steps) gave 59.0\% exact match and 100\% name F1 on BitAgent held-out data.
The same model scores 37.3\% on Nemotron and 5.3\% on the multi-call set. Most of the spread between datasets reflects how hard each evaluation set is.

\begin{table}[t]
\centering\small
\caption{Function calling on held-out data (needle-style metrics, \%). ``Grammar'' is grammar-constrained decoding.}\label{tab:tools}
\begin{tabular}{@{}l l r r r r r@{}}
\toprule
Model & Eval set & JSON & Name F1 & Args & Param.\ halluc.\ $\downarrow$ & Exact \\
\midrule
Nemotron SFT, greedy & Nemotron (400) & 93.2 & 82.2 & 51.8 & 3.8 & 38.5 \\
\quad + grammar & Nemotron (400) & -- & -- & -- & 0.0 & 38.5 \\
\quad + weighted loss + grammar & Nemotron (400) & 94.3 & 83.4 & 52.9 & 0.6 & 39.5 \\
Combined data + grammar & BitAgent & -- & 100.0 & -- & 0.0 & 59.0 \\
Combined data + grammar & Nemotron & -- & -- & -- & -- & 37.3 \\
\bottomrule
\end{tabular}
\end{table}

\paragraph{BFCL and the call/abstain tension.}
On BFCL v3~\citep{patil2025bfcl} (AST scoring, 150 examples per category, grammar-constrained), the combined model scores:
\begin{itemize}
\item 22.7 on \emph{simple} and 29.3 on \emph{multiple};
\item about 0 on \emph{parallel}, because the training data contains only the first call of each conversation;
\item 77.8 on \emph{relevance}, but only 15.3 on \emph{irrelevance}.
\end{itemize}
Adding multi-turn trajectory data taught abstention: irrelevance rose to 90.7 with a 21\% trajectory mix and to 95.3 with trajectories only.
However, it also collapsed must-call accuracy (simple 22.7 to 0.7 to 0.0). At 500M parameters, rigid calling and abstention trade off against each other.
Our explanation is that the prose summaries in the trajectories teach the model that an assistant turn is usually text. They do not teach it to tell matching tools from mismatched ones.
For deployment we therefore use a forced-call policy whenever tools are supplied. The model's reply is seeded with \code{[}, the predicted name is snapped to the closest valid name, and the JSON is repaired if needed.

\subsection{Computer use from accessibility trees}\label{sec:cua}
Our GUI agent reads text only: it never sees pixels.
The screen is serialized from the Windows UI-Automation tree~\citep{microsoft_uia} as lines of the form \code{[ControlType] "Name" (state)}.
The model emits one of 12 actions as JSON: click, double-click, right-click, hover, type, key press, scroll, drag, select, set value, open app and wait. The action names its target element by name and control type rather than by coordinates.
Because a 2,048-token context could not hold raw trees (58\% of 100-element trees overflowed), we cap each tree at 45 elements, interactive controls first, identically at training and inference.
We built the training data in three generations, with Gemini~\citep{team2023gemini} as the instruction writer and our own validators as filters:
\begin{enumerate}
\item synthetic screens of 30 application archetypes (7,200 training rows);
\item instructions written over 28 harvested real Windows trees (1,810 rows);
\item multi-step goals expanded into \code{GOAL / DONE / NEXT} rows, with a 13th \code{finish} action (1,472 rows).
\end{enumerate}
Every generated action is checked against the tree: elements that are not present are rejected, and wrong control types are corrected.
Table~\ref{tab:cua} gives held-out results for the released model.
Parameter hallucination is zero by construction of the data, and most exact-match misses are ambiguous values (``give it a minute'' $\to$ \code{wait(60)} vs.\ a reference of 7).
On long goals, per-step errors compound, and the agent has no mechanism to recover from a wrong step.
In the omni model (joint7 onward) the same tool skills also work from speech. For example, a spoken weather question is answered by calling the tool, observing the result and summarizing it in one checkpoint.

\begin{table}[t]
\centering\small
\caption{Held-out computer-use results of HobbyLM-Computer-Use (\%).}\label{tab:cua}
\begin{tabular}{@{}l r r r r r@{}}
\toprule
Split & JSON & Name F1 & Value acc. & Exact & Param.\ halluc. \\
\midrule
Planning (multi-step goals) & 96.5 & 94.7 & -- & 82.6 & 0.0 \\
Grounding (real application trees) & ${\sim}96$ & 95.5 & 91 & 78.4 & 0.0 \\
Grounding (synthetic screens) & 100 & 90.7 & 88.6 & 72.5 & 0.0 \\
\bottomrule
\end{tabular}
\end{table}

% ===========================================================================
\section{HobbyLM-1B: adopting the architecture}\label{sec:1b}
\emph{Work by Hariharan M.\,S. and Prabhurajhan R. The architecture is the \code{1B} preset of Section~\ref{sec:arch}, designed by Harish Santhanalakshmi Ganesan}

HobbyLM-1B was built in a separate, private code base (revision \code{86dbb90}). The public repository \url{https://github.com/fuellabs-ai/hobbylm-1b} documents the released models; it does not contain the training code.
Table~\ref{tab:adopt} separates what was adopted from what was added.
Several analyses in this section only make sense in light of specific architectural choices: router precision follows from sigmoid routing with a selection bias at 64 experts, and the micro-batch trade-off follows from per-forward bias updates. Where that is the case, the text links back to Section~\ref{sec:arch}.

\begin{table}[t]
\centering\small
\caption{HobbyLM-1B: adopted from the architecture vs.\ added by the 1B build.}\label{tab:adopt}
\begin{tabularx}{\linewidth}{@{}l Y@{}}
\toprule
Adopted unchanged & 1B preset (Table~\ref{tab:presets}); sigmoid top-8-of-64 routing with aux-loss-free bias ($u=10^{-3}$, frozen at 95\%); shared expert; QK-norm; GQA; router z-loss $10^{-3}$, balance loss $10^{-3}$, final-logit z-loss $10^{-4}$; chunked cross-entropy; fp32 router; Muon/AdamW split and $\sqrt{B}$ learning-rate scaling; warmup--stable--decay schedule; shard format \\
\midrule
Added by the 1B build & 100B-token curated mixture; tokenizer study; stratified shard ordering; anneal phase on high-quality data; 4$\times$H200 DDP training configuration; checkpoint-backup automation; evaluation; data-order and router-precision analyses; instruction tuning; macOS computer-use adapter \\
\bottomrule
\end{tabularx}
\end{table}

\subsection{Data}

\paragraph{Mixture.}
HobbyLM-1B was trained on a schedule of 100B tokens drawn from five sources (Table~\ref{tab:data1b}):
\begin{itemize}
\item FineWeb-Edu~\citep{penedo2024fineweb}, as pre-tokenized GPT-2 shards~\citep{karpathy_fineweb_edu_shards};
\item DCLM-baseline~\citep{li2024dclm};
\item \code{codeparrot-clean}~\citep{codeparrot}, used as a fallback because \code{bigcode/starcoderdata} was gated for the team's access token at the time; it is mostly Python;
\item FineMath-4plus~\citep{allal2025smollm2};
\item an anneal set of Cosmopedia-v2 from the SmolLM-Corpus~\citep{benallal2024smollmcorpus} and FineMath-4plus.
\end{itemize}
Validation uses one held-out FineWeb shard~\citep{kjj0_fineweb10b}.

\begin{table}[t]
\centering\small
\caption{HobbyLM-1B training data.}\label{tab:data1b}
\begin{tabular}{@{}l l r r r l@{}}
\toprule
Source & Dataset & Shards & Tokens & Share & Phase \\
\midrule
Educational web & FineWeb-Edu (shards 000001--000600) & 600 & 60B & 60\% & Main \\
Filtered web & DCLM-baseline-1.0 & 150 & 15B & 15\% & Main \\
Code & codeparrot-clean & 100 & 10B & 10\% & Main \\
Math & FineMath-4plus & 100 & 10B & 10\% & Main \\
Anneal & Cosmopedia-v2 (3.5B) + FineMath-4plus (1.5B) & 50 & 5B & 5\% & Anneal \\
\bottomrule
\end{tabular}
\end{table}

\paragraph{Tokenizer.}
The team kept the GPT-2 tokenizer, which preserved direct comparability with the 130M experiments.

\paragraph{Shards and ordering.}
All data was pre-tokenized into 100M-token shards in the format of Section~\ref{sec:arch}: a 1,024-byte header (magic 20240520, version, token count) followed by uint16 token IDs, with documents separated by the end-of-text token.
Documents of 80 characters or fewer were dropped.
Shards are consumed whole, so their order determines what the model sees over time.
The team uses a seeded, stratified order that interleaves the four main-phase families in proportion to their shard counts. They verified the order on all 950 main-phase shards before launch (Table~\ref{tab:thirds}).
Section~\ref{sec:order} shows why this matters.

\begin{table}[t]
\centering\small
\caption{Family mix in each third of the 1B main-phase shard order.}\label{tab:thirds}
\begin{tabular}{@{}l r r r r r@{}}
\toprule
Third & Shards & Edu & DCLM & Code & Math \\
\midrule
1 & 317 & 63.09\% & 15.77\% & 10.73\% & 10.41\% \\
2 & 317 & 63.09\% & 15.77\% & 10.41\% & 10.73\% \\
3 & 316 & 63.29\% & 15.82\% & 10.44\% & 10.44\% \\
\midrule
Target & & 63.2\% & 15.8\% & 10.5\% & 10.5\% \\
\bottomrule
\end{tabular}
\end{table}

\paragraph{Tokens consumed vs.\ unique.}
The schedule consumes $95{,}367$ steps $\times\,1{,}048{,}576$ tokens $\approx 100.0$B tokens.
\begin{itemize}
\item The main phase (81,062 steps, ${\approx}85.0$B tokens) draws from a 95B-token pool.
\item The anneal (14,305 steps, ${\approx}15.0$B tokens) makes about three passes over the 5B anneal set.
\end{itemize}
The anneal's FineMath portion streams from the start of the same split used for the main math shards, so the two likely overlap by about 1.5B tokens. The team therefore estimates 88.5--90B unique tokens.

\subsection{Training}
Table~\ref{tab:config1b} gives the configuration.
The learning-rate schedule spans all 95,367 steps.
The main phase stops at step 81,062. The anneal resumes from that checkpoint with the same optimizer state and schedule position, and changes only the data.
At the resume, the Muon learning rate was 0.0175, confirming that the cooldown continued rather than restarted.

\begin{table}[t]
\centering\small
\caption{HobbyLM-1B training configuration.}\label{tab:config1b}
\begin{tabularx}{\linewidth}{@{}l Y@{}}
\toprule
Hardware & 4$\times$ NVIDIA H200 SXM, one node \\
Parallelism & PyTorch DDP via \code{torchrun}; gradient all-reduce only on the last micro-step \\
Sequence / micro-batch / accumulation & 1,024 / 32 per GPU / 8 \\
Tokens per optimizer step & 1,048,576 \\
Precision & bf16 autocast, fp32 master weights, TF32 matmuls, fp32 router logits \\
Loss & chunked cross-entropy; final-logit z-loss $10^{-4}$; router z-loss $10^{-3}$; balance loss $10^{-3}$ \\
Compilation & whole-model \code{torch.compile} \\
Peak learning rate & Muon 0.04, AdamW $6\times10^{-4}$ (base 0.02 / $3\times10^{-4}$, scaled by 2) \\
Muon / AdamW & Nesterov 0.95, weight decay 0.1, 5 Newton--Schulz steps / $\beta=(0.9,0.95)$, weight decay 0.1 \\
Schedule & 100-step warmup; constant; linear decay over the last 40\% to $0.1\times$ peak \\
Gradient clipping & 1.0 \\
Expert balancing & aux-loss-free bias, $u=10^{-3}$, frozen at step 90,598 (95\%) \\
Seed & 1337 \\
\bottomrule
\end{tabularx}
\end{table}

\paragraph{Compute.}
Main pretraining took about 68\,h and the anneal about 8\,h of wall-clock time on four GPUs: about 76\,h in total, or about 304 GPU-hours.
That total excludes the 130M experiments, failed launches, sanity checks, evaluation and post-training.
Billing for the instance, including data staging and checks, was about \$1,167.

\paragraph{Engineering issues.}
\begin{itemize}
\item \textbf{Pre-step hang.} On the first provider, every launch hung before step~0. The model code contained two nested \code{torch.compile} decorators inside the custom operators of the experimental FP8 output head (Section~\ref{sec:fusedce}). The head was not enabled in the 1B configuration. After both decorators were removed, compiled runs worked on a second provider. Because the environment also changed, the exact cause was not established.
\item \textbf{Checkpoint storage.} Checkpoints (weights plus optimizer state, about 8.5\,GB each) filled local storage after an upload quota was hit. The backup process now deletes local checkpoints only after a confirmed upload.
\end{itemize}

\subsection{Results}
We report the builders' release evaluation of the final base model (Table~\ref{tab:results1b}): lm-eval 0.4.13, fp32, 0-shot, all examples of each split, with \code{acc\_norm} for HellaSwag, ARC, PIQA and OpenBookQA and \code{acc} for WinoGrande and BoolQ.
These are the same metric choices as Table~\ref{tab:lmeval}, which used lm-eval 0.4.9.1.
The seven-task average is 47.54. Like Table~\ref{tab:lmeval}, it mixes \code{acc} and \code{acc\_norm}; the builders' model card reports no average for that reason.

\begin{table}[!htbp]
\centering\small
\caption{HobbyLM-1B base, 0-shot release evaluation (lm-eval 0.4.13, fp32). $\pm$ is the lm-eval standard error.}\label{tab:results1b}
\begin{tabular}{@{}l l r@{}}
\toprule
Task & Metric & Final (step 95,367) \\
\midrule
HellaSwag & acc\_norm & 43.66 $\pm$ 0.49 \\
PIQA & acc\_norm & 67.85 $\pm$ 1.09 \\
WinoGrande & acc & 52.41 $\pm$ 1.40 \\
ARC-Easy & acc\_norm & 54.76 $\pm$ 1.02 \\
OpenBookQA & acc\_norm & 35.00 $\pm$ 2.14 \\
ARC-Challenge & acc\_norm & 29.27 $\pm$ 1.33 \\
BoolQ & acc & 49.82 $\pm$ 0.87 \\
\midrule
\textbf{Average} & & \textbf{47.54} \\
\bottomrule
\end{tabular}

\smallskip
\parbox{0.8\linewidth}{\footnotesize One BoolQ document (id 3154) exceeds the base model's 1,024-token context and was left-truncated; without it, BoolQ is 49.83. Standard errors and run details are in the base model card.}
\end{table}

\subsection{Analysis}

\paragraph{Data order at 130M.}\label{sec:order}
Before the 1B run, the team tested the mixture on the 130M preset (140.0M total, 62.1M active) with 5B tokens: 4,050 main-phase and 720 anneal steps (Table~\ref{tab:order}).
Their first loader expanded each source's file pattern in order. Even with shuffling enabled, it chose among sources rather than across all shards. In practice it consumed educational web, then DCLM, then code, and never reached math.
Validation loss rose from 3.74 to 5.13 at the end of that run, and the curated mixture scored below the FineWeb baseline even after a rescue anneal.
With stratified ordering, the same data and model matched the baseline (43.34 vs 42.97, single runs) with half the tokens.
Shard order is effectively a training hyperparameter whenever sources are stored separately. A cheap check catches the failure before launch: print the source of the first shards and the mix in each third (as in Table~\ref{tab:thirds}).
The baseline's average (42.97) equals that of the 130M FineWeb-10B model in Table~\ref{tab:lmeval}.

\begin{table}[t]
\centering\small
\caption{130M data-order experiments, 7-task average; the baseline is the 130M FineWeb-10B model of Table~\ref{tab:lmeval}.}\label{tab:order}
\begin{tabular}{@{}l l r r@{}}
\toprule
Run & Data & Tokens & Average \\
\midrule
Baseline & FineWeb & 10B & 42.97 \\
Sequential loader & curated mix, consumed source by source & 5B & 40.70 \\
Stratified order & curated mix, interleaved & 5B & 43.34 \\
\bottomrule
\end{tabular}
\end{table}

\paragraph{Router precision.}
The released implementation uses fp32 routing. A saved precision diagnostic showed changes in routing statistics under bf16, but did not establish a benchmark degradation. These observations do not demonstrate that fp32 routing is universally necessary.
Quantized GGUF builds keep the router in fp32. On seven fixed prompts, Q8\_0 reproduced the recorded Hugging Face fp32 outputs on all seven, and Q4\_K\_M on six; F32 GGUF also matched six. Relative to F32 GGUF, perplexity increased by approximately 0.02\% for Q8\_0 and 1.1\% for Q4\_K\_M on a corpus composed of evaluation prompts and model-generated answers. These are limited compatibility and quantization checks, not a broad capability evaluation.

\subsection{HobbyLM-1B-Instruct}\label{sec:1b-instruct}
Hariharan M.\,S. and Prabhurajhan R. also fine-tuned the base model into HobbyLM-1B-Instruct, which follows simple instructions, rewrites and extracts text, and makes single function calls in a Python-list format.
The released checkpoint is not fine-tuned directly from the 1,024-token base (Appendix~\ref{app:sft1b}, Table~\ref{tab:lineage1b}).
Its lineage first extends the base to 2,048 tokens over 550 steps, then applies a 100-step function-calling pilot, followed by a 150-step extension to 4,096 tokens with plain RoPE ($\theta=10^4$, no RoPE scaling) that starts from the pilot's weights.
The resulting checkpoint (R2) failed its recorded 4K retrieval evaluation, so training at sequence length 4,096 should not be read as demonstrated retrieval across that context.

R2 initializes a four-stage SFT path: 500 recovery steps over 15 data buckets; 250 continuation steps over 18 buckets; 2,000 further steps over 19 buckets to step 2,750; and a final 700-step Broad-Instruct continuation over 19 buckets, four of them rebuilt.
The SFT step counter starts at the recovery stage, so step 3,450 counts 3,450 SFT optimizer steps along the selected branch, excluding the 800 extension and pilot steps.
The preceding continuation run also produced checkpoints up to step 4,750; these are not ancestors of the release.
Checkpoint metadata and the inspected trainer are consistent with optimizer, token-counter and sampler state being restored between SFT stages, with selected buckets reinitialized during data migrations.
The final stage's launch command and the exact trainer file executed remotely were not preserved.
The released weights are identical to the raw step-3,450 checkpoint (280 of 280 tensors match).

Public SFT sources include SmolTalk~\citep{allal2025smollm2} subsets for dialogue, rewriting, constraints and summarization; Nemotron-SFT-Agentic-v2~\citep{nvidia_nemotron_sft_agentic_v2}; and callable examples derived from Glaive function-calling v2~\citep{glaive_fc_v2}, ToolACE~\citep{liu2025toolace} and xLAM-60k~\citep{liu2024apigen}.
Other buckets contain internally generated math, Python, grounded-answering, instruction-following, structured-output and tool-use examples; the upstream sources of the semantic-abstention bucket are not fully traced.
Table~\ref{tab:sftmix1b} lists the prepared final-stage datasets and their configured sampling shares; these describe the mixture, not the number of distinct examples consumed.
Training uses 4,096-token sequences and 256 sequences per optimizer step. The step-3,450 checkpoint records about 3.62B processed token positions and 78.1M supervised tokens (SFT plus replay) over the selected path; neither is a count of unique tokens.
The optimizer, objective and loss-masking details are in Appendix~\ref{app:sft1b}.
The model has no preference or safety tuning.
It is released with GGUF (Q4\_K\_M, Q8\_0, F32) and MLX builds (Appendix~\ref{app:artifacts}).

Table~\ref{tab:instruct1b} compares the two models under the same evaluation (lm-eval 0.4.13, fp32, 0-shot, all examples).
Changes are observed differences only; we ran no significance test.
CommonsenseQA and MMLU are at chance level for both models (20\% and 25\%).
On IFEval (all 541 prompts, official lm-eval 0.4.13 scoring, the model's chat template, greedy decoding, a 2,048-token response budget), HobbyLM-1B-Instruct scores 22.92 prompt-level strict, 37.77 instruction-level strict, 24.95 prompt-level loose and 40.53 instruction-level loose.
135 of 541 responses hit the 2,048-token limit. Two prompts were flagged by an overlap check against the final-stage SFT mixture; earlier stages and pretraining were not checked.

\begin{table}[!htbp]
\centering\small
\caption{HobbyLM-1B base vs.\ Instruct, 0-shot (lm-eval 0.4.13, fp32, \%). The average is over the seven tasks of Table~\ref{tab:lmeval}.}\label{tab:instruct1b}
\begin{tabular}{@{}l l r r@{}}
\toprule
Benchmark & Metric & Base & Instruct \\
\midrule
HellaSwag & acc\_norm & 43.66 & 46.64 \\
ARC-Easy & acc\_norm & 54.76 & 54.12 \\
ARC-Challenge & acc\_norm & 29.27 & 28.41 \\
PIQA & acc\_norm & 67.85 & 68.88 \\
WinoGrande & acc & 52.41 & 50.75 \\
OpenBookQA & acc\_norm & 35.00 & 36.20 \\
BoolQ & acc & 49.82 & 53.36 \\
\midrule
\emph{7-task average} & & \emph{47.54} & \emph{48.34} \\
\midrule
SIQA & acc & 40.89 & 40.33 \\
CommonsenseQA & acc & 19.98 & 18.18 \\
SciQ & acc / acc\_norm & 82.90 / 75.80 & 82.70 / 77.10 \\
MMLU (0-shot) & acc & 25.31 & 24.78 \\
\bottomrule
\end{tabular}
\end{table}

\paragraph{CPU inference.}
The GGUF builds keep the router and expert-bias tensors in F32. Like the 500M export (Section~\ref{sec:infer}), they use llama.cpp's \code{bailingmoe2} layout with the one-line decoupled-head-dimension patch.
Stock Ollama v0.35.1 fails to load them, so the release ships patched llama.cpp and Ollama builds and a double-click Windows chat bundle.
On one AMD Ryzen 9 5900HX, the Q4\_K\_M Instruct model generates about 14 tokens/s with about 1.3\,GB peak memory.

\paragraph{Computer use.}
Prabhurajhan R. also trained HobbyLM-1B Computer Use v0.1, a LoRA adapter for HobbyLM-1B-Instruct that turns plain-English requests into Mac actions offline. It is released with its app, harness and evaluation suite (Appendix~\ref{app:artifacts}); we do not evaluate it in this paper.

\FloatBarrier
% ===========================================================================
\section{HobbyLM-Image: text-to-image generation}\label{sec:image}
\emph{Work by Harish Santhanalakshmi Ganesan}

HobbyLM-Image is a separate 333M-parameter diffusion transformer (DiT)~\citep{peebles2023dit} trained with a flow-matching objective~\citep{lipman2023flowmatching,liu2023rectflow}. It generates $1024\times1024$ images.
Training the main lineage cost about \$300 of GPU time, about \$445 including all experiments.
That budget is possible because the model works in a deep-compression latent space rather than a conventional VAE latent.

\subsection{Architecture}
\paragraph{Latent space.}
We use the frozen DC-AE f32c32 autoencoder released with SANA~1.1~\citep{chen2025dcae,xie2025sana,mitdcae_hf}. It compresses $32\times$ per spatial dimension, so a $1024^2$ image becomes a $32\times32\times32$ latent.
Because the latent is already so compressed, the DiT uses patch size 1: every latent pixel is one token.
After applying DC-AE's published scaling factor, we further normalize latents to unit variance using their empirical standard deviation on the training cache. We found that this stabilized training.
In our check, DC-AE round-trip reconstruction was about 26\,dB PSNR at 512px.

\paragraph{Two-panel canvas.}
The model was designed to both generate and edit images. Its input is a two-panel canvas $[\,\text{noisy target}\mid\text{source}\,]$ concatenated along width, plus two mask channels (34 input channels).
For generation the source panel is blank. For editing it holds the source latent.
A learned left/right panel embedding and a task embedding tell the model which case applies. The loss is taken on the left (target) panel only.
At 512px the canvas is 512 tokens. At 1024px it is 2,048 tokens.

\paragraph{Blocks.}
The DiT has 16 layers of width 1024 with 16 heads and MLP ratio 3 (GELU-tanh). Each block applies:
\begin{enumerate}
\item AdaLN-Zero~\citep{peebles2023dit} self-attention, conditioned on the timestep and task embeddings;
\item ungated cross-attention to the text features;
\item an AdaLN-Zero MLP.
\end{enumerate}
Positions are learned absolute embeddings.
Text conditioning uses the frozen CLIP ViT-L/14 text encoder~\citep{radford2021clip} (prompts padded or truncated to 64 tokens, width 768).
The original plan conditioned on our own 500M VLM. We replaced it after the VLM failed explicit grounding probes (Section~\ref{sec:mm}; joint12 scored near the language-prior baseline on strict GQA and VQAv2).
As in the language model, we apply RMSNorm to $Q$ and $K$ per head before the dot product, in both self- and cross-attention.
Without QK-norm the DiT diverged repeatedly. Lowering the learning rate only delayed the divergence, and adding QK-norm removed it.

\subsection{Training}
For a clean latent $z_1$, noise $z_0\sim\mathcal N(0,I)$ and $t$ drawn from a logit-normal distribution~\citep{esser2024sd3}, the model $v_\theta$ regresses the rectified-flow velocity:
\begin{equation}
z_t = (1-t)\,z_0 + t\,z_1,\qquad
\mathcal L = \mathbb E\big[\,w\,\|v_\theta(z_t,t,c) - (z_1 - z_0)\|^2\big].
\end{equation}
For edits, the weight $w$ is $1+1.5\,m$, which up-weights the masked foreground $m$. For generation, $w=1$.
To prevent occasional metastable loss jumps, we clamp the per-sample loss at 4.0, skip non-finite batches, clip gradients at 1.0 and keep an EMA of the weights (decay 0.999).
For classifier-free guidance~\citep{ho2022cfg}, the text condition is dropped 10\% of the time, and masks are dropped 50\% of the time so that mask-free inference stays in distribution.
We use AdamW with $\beta=(0.9,0.95)$ and weight decay 0.02, linear warmup, and a cosine decay to 10\% of the peak learning rate. The peak is $5\times10^{-5}$ from scratch and $2\times10^{-5}$ for continued training.
On a B200 the 512px DiT trains at about 887 images/s, which made a 100M-image-view run cost about \$217.

\subsection{Data curriculum}
Each stage was triggered by a specific failure we saw in samples (Table~\ref{tab:imgdata}).
The 512px base model learned from CC3M and CC12M~\citep{sharma2018cc3m,changpinyo2021cc12m} plus editing pairs from OmniEdit~\citep{wei2025omniedit} and MagicBrush~\citep{zhang2023magicbrush}. These web-alt-text images taught it stock-photo \emph{watermarks}.
Continued training on ImageNet~\citep{deng2009imagenet} objects and watermark-free Midjourney-v6 recaptions~\citep{photoroom_mj} removed the watermarks. UltraEdit~\citep{zhao2024ultraedit} and OmniEdit edits were replayed alongside to retain editing.
The Midjourney-heavy mix in turn made objects look painterly and melted.
At 1024px we fine-tuned on FLUX-Reason-6M~\citep{fang2025fluxreason}, images generated by FLUX.1-dev~\citep{bfl2024flux} that are literal and accurate. This corrected object geometry. A caption-filtered subset of images with people and a small amount of Midjourney data were added for human subjects and aesthetics.
When we moved from 512px to 1024px, we interpolated the learned position embeddings bicubically.
The final checkpoint, \code{1024flux4}, reached a flow-matching loss of 0.76, the lowest in the lineage. The loss was still decreasing when training stopped.

\begin{table}[t]
\centering\small
\caption{HobbyLM-Image training curriculum. Data counts are the configured per-source quotas; step counts are the defaults of the corresponding training actions.}\label{tab:imgdata}
\begin{tabularx}{\linewidth}{@{}l l Y r@{}}
\toprule
Stage & Res. & Data & Steps \\
\midrule
Base & 512 & CC3M (140K), CC12M (320K), OmniEdit (40K), MagicBrush (2K) & -- \\
Continued & 512 & ImageNet (50K), Midjourney-v6 recap (1M), UltraEdit (100K), OmniEdit (100K); 20--25\% edit replay & 45K + 120K \\
Sharpness & 1024 & continued mix at 1024px & 20K \\
Accuracy & 1024 & FLUX-Reason-6M (500K + 150K people), Midjourney (50K) & up to 100K per run \\
\bottomrule
\end{tabularx}
\end{table}

\subsection{Sampling and results}
\begin{figure}[t]
\centering
\includegraphics[width=0.49\linewidth]{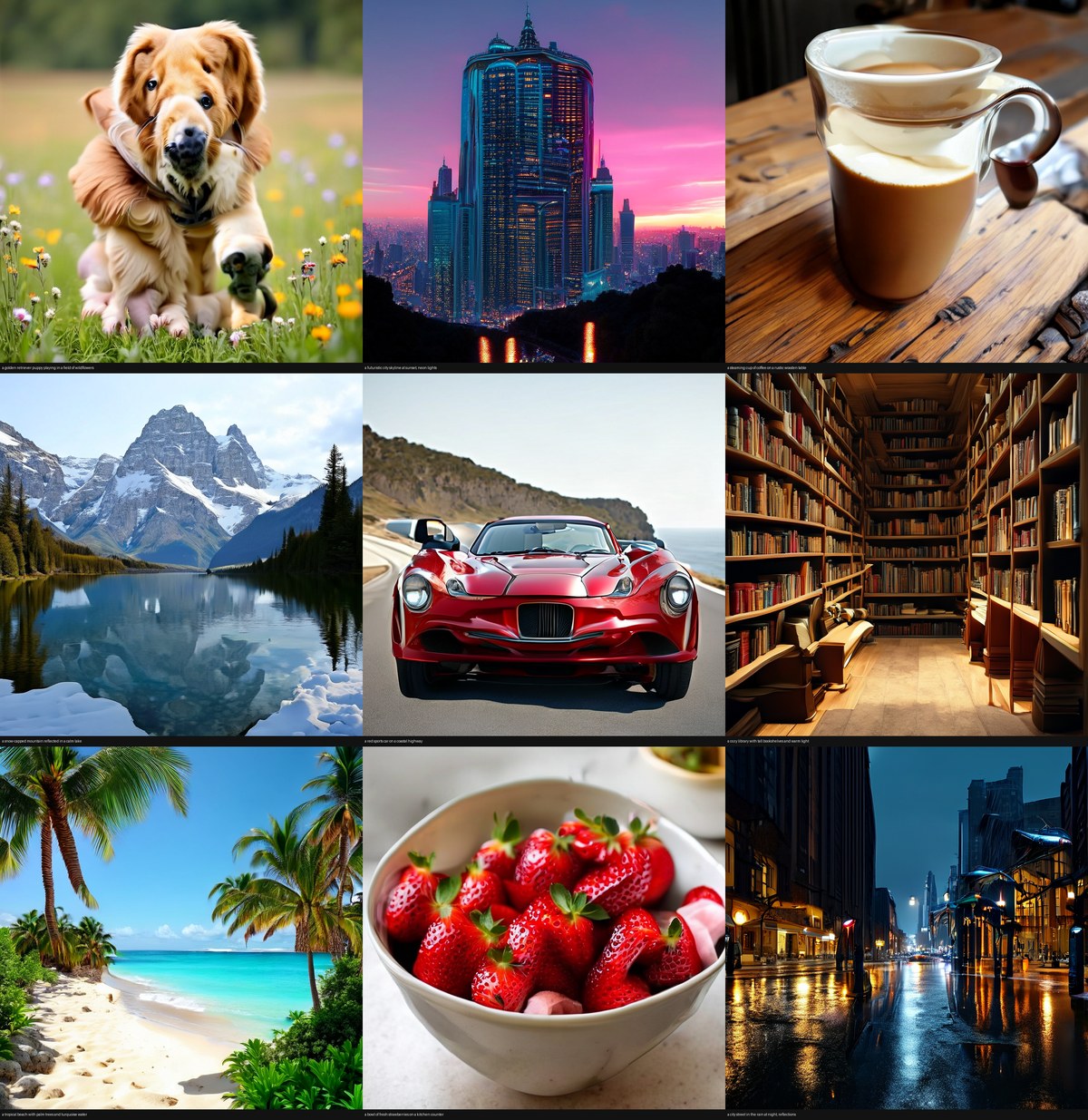}\hfill
\includegraphics[width=0.49\linewidth]{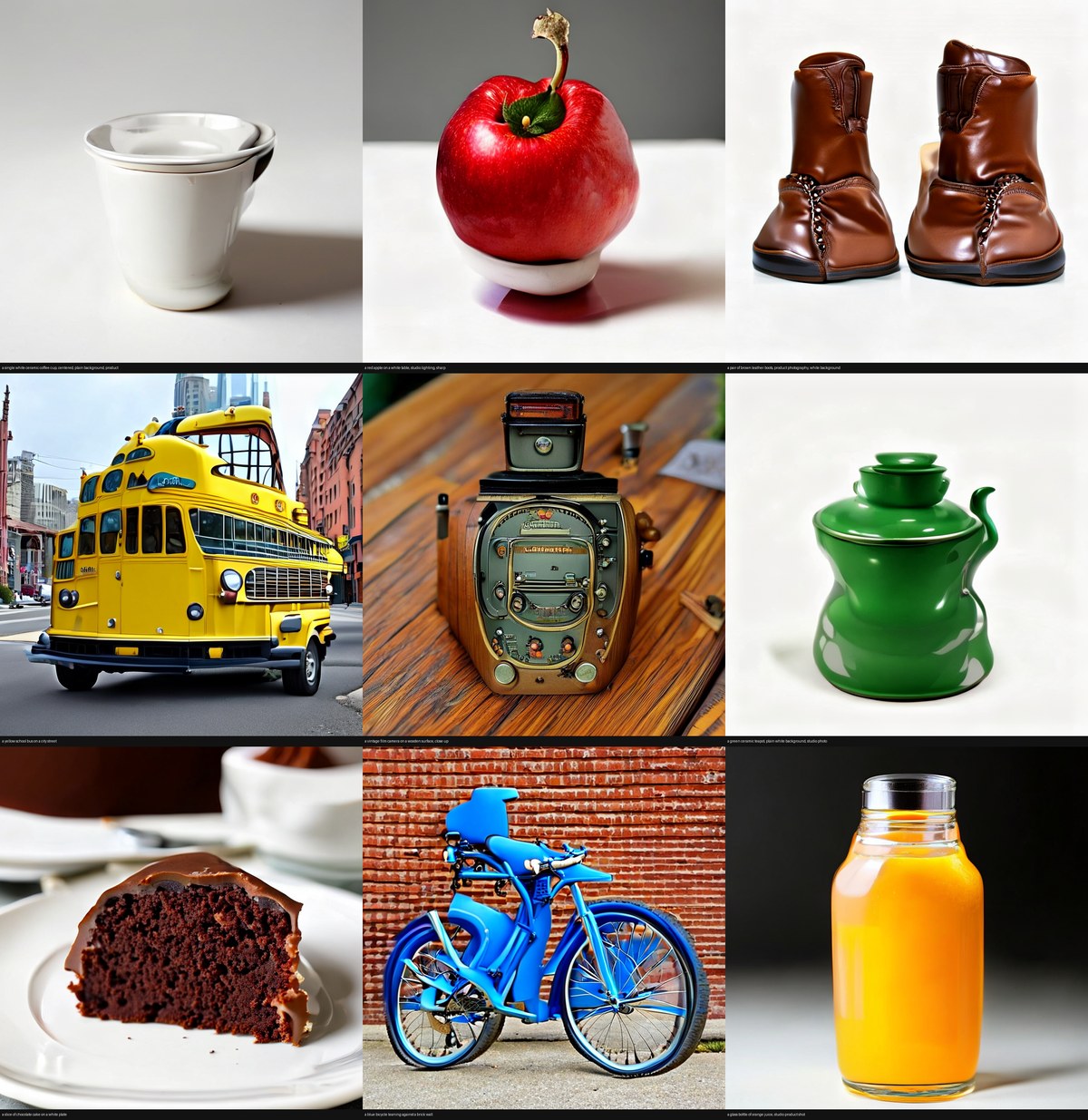}
\caption{$1024\times1024$ samples from HobbyLM-Image (final checkpoint, CFG $\approx5$, about 100 Euler steps), downscaled. Left: scene prompts. Right: object prompts. The object panel shows typical failures in fine structure (bus, bicycle).}\label{fig:samples}
\end{figure}

Sampling integrates the learned velocity field with an Euler solver from noise to data, using about 100 steps and a guidance scale around 5:
\begin{equation}
z \leftarrow z + \tfrac{1}{n}\big(v_\theta(z,t,\varnothing) + \gamma\,[v_\theta(z,t,c) - v_\theta(z,t,\varnothing)]\big).
\end{equation}
Figure~\ref{fig:samples} shows samples from the final checkpoint.
Single objects and scenes are recognizable and well lit.
Fine structure is often wrong (for example the bus and bicycle in the right panel of Figure~\ref{fig:samples}), and people, hands and multi-person scenes are soft.
We did not compute FID or GenEval.
A 512px sibling checkpoint performs instruction-based editing through the source panel.

\subsection{CPU implementation}
We also ported the full pipeline to \code{hobby-rs} with no ML framework dependency. The port consists of:
\begin{itemize}
\item a safetensors reader;
\item the CLIP text encoder and its BPE tokenizer;
\item the DiT and the CFG flow-matching sampler;
\item the DC-AE decoder (residual blocks, linear attention and GLU-MBConv blocks, following the diffusers reference);
\item a PNG writer.
\end{itemize}
A staged verifier compares every stage with PyTorch. The CLIP features match to a maximum absolute difference of $1.7\times10^{-5}$, and the DiT velocities to about $10^{-5}$.
Generating a $1024^2$ image takes minutes on a laptop CPU. We accept this cost in exchange for having no framework dependency.

% ===========================================================================
\section{Limitations}\label{sec:limits}
\begin{itemize}
\item \textbf{Token budget.} The 500M model saw 40B tokens and the 1B model 100B, one to three orders of magnitude fewer than the strongest models of similar size. Every downstream variant inherits this ceiling: the models are fluent but hallucinate.
\item \textbf{Single runs.} Every model was trained once, with one seed. The ablations are single runs at 130M and 3,000 steps, so differences of about 0.01 are within noise.
\item \textbf{Multimodal grounding.} POPE shows the stage-2 VLM answering ``yes'' to every probe. Strict GQA/VQAv2 scores of the final omni model are near language-prior baselines. The stage-2 GQA and VQAv2 numbers use 2K-question subsets.
\item \textbf{Agents.} Function calling is strong on single-call distributions similar to training data but weak on parallel calls. Computer-use errors compound over long goals. The agents act on a real desktop; our harnesses limit execution to a safe action set or a dry-run mode by default.
\item \textbf{1B comparability.} Same-harness baselines for the 1B model are pending. HobbyLM-1B base has a 1,024-token context. HobbyLM-1B-Instruct is configured for 4,096 tokens, but long-context retrieval is unverified, and its parent checkpoint failed a 4K retrieval test. The HobbyLM-1B training and data-preparation code is not public. No model in the family has been evaluated for safety or bias, and none has been safety-tuned.
\item \textbf{Image model.} We report no FID or GenEval scores. The training images include synthetic images from other generators, which may carry over their biases.
\end{itemize}

% ===========================================================================
\section{Conclusion}
A single carefully ablated sparse-MoE architecture supported an unusually wide range of models on a small budget: base, chat, diffusion, omni-modal, function-calling and computer-use models at 500M, an independently built 1B model, and the same stability recipe (QK-norm) in a separate image generator.
Three lessons carried across the whole project:
\begin{itemize}
\item Get the normalization and routing math right first.
\item Treat data order and decoding as hyperparameters.
\item Measure on the hardware you care about, because memory- versus compute-bound regimes can reverse conclusions.
\end{itemize}

\section*{Acknowledgments}
We thank Jarvis Labs for GPU credits toward the HobbyLM-1B training run.

\paragraph{AI assistance.} For the HobbyLM-1B work, the authors used Anthropic's Claude through Claude Code and OpenAI's Codex to assist with writing and debugging data-preparation, training, evaluation, conversion and release scripts; inspecting and summarizing saved experimental artifacts; and drafting and revising documentation and portions of the manuscript. The authors take responsibility for the manuscript and its claims.

\bibliographystyle{plainnat}
\bibliography{references}

% ===========================================================================
\FloatBarrier
\clearpage
\appendix

\section{Artifacts}\label{app:artifacts}
Table~\ref{tab:artifacts} lists every released artifact.

\begin{table}[h]
\centering\small
\caption{Released artifacts. Hugging Face repositories are under \url{https://huggingface.co/}.}\label{tab:artifacts}
\footnotesize
\begin{tabularx}{\linewidth}{@{}>{\raggedright\arraybackslash}p{2.9cm} >{\raggedright\arraybackslash}p{6.4cm} Y@{}}
\toprule
Artifact & Location & Description \\
\midrule
HobbyLM-Base & \code{rootxhacker/HobbyLM-Base} & 500M base model, 8k context (Section~\ref{sec:500m}) \\
HobbyLM-Chat & \code{rootxhacker/HobbyLM-Chat} & SmolTalk SFT and UltraFeedback DPO \\
HobbyLM-Computer-Use & \code{rootxhacker/HobbyLM-Computer-Use} & function calling and accessibility-tree agent (Section~\ref{sec:cua}) \\
HobbyLM-Omni & \code{rootxhacker/HobbyLM-Omni} & joint12 omni checkpoint with vision and speech projectors (Section~\ref{sec:mm}) \\
HobbyLM-Diffusion & \code{rootxhacker/HobbyLM-Diffusion} & masked-diffusion chat model (Section~\ref{sec:diffusion}) \\
HobbyLM-SAE & \code{rootxhacker/HobbyLM-SAE} & top-$k$ SAE on layer 8 of HobbyLM-Base \\
HobbyLM-gguf & \code{rootxhacker/HobbyLM-gguf} & GGUF builds of the language models (\code{hobbylm} architecture) \\
HobbyLM-Image & \code{rootxhacker/HobbyLM-Image} & 333M text-to-image DiT (Section~\ref{sec:image}) \\
HobbyLM-Playground & \code{spaces/rootxhacker/HobbyLM-Playground} & interactive demo \\
HobbyLM code & \code{github.com/harishsg993010/HobbyLM} & training, evaluation, export, \code{hobby-rs}, \code{hobby-chat} \\
\midrule
HobbyLM-1B & \code{harims95/hobbylm-1B} & 1B base model (Section~\ref{sec:1b}) \\
HobbyLM-1B-Instruct & \code{harims95/hobbylm-1B-instruct} & instruction-tuned 1B (Section~\ref{sec:1b-instruct}) \\
HobbyLM-1B MLX & \code{harims95/hobbylm-1B-mlx} & Instruct for Apple silicon (MLX) \\
HobbyLM-1B GGUF & \code{harims95/hobbylm-1B-gguf} & Instruct in Q4\_K\_M, Q8\_0 and F32; base in F32. Needs the patched llama.cpp from the v1.0.0 release; stock Ollama does not load these files \\
HobbyLM-1B Computer Use v0.1 & \code{Hobbylm/hobbylm-1B-computer-use} & LoRA adapter for HobbyLM-1B-Instruct, Mac actions \\
HobbyLM-1B chat & \code{spaces/harims95/hobbylm-1b-chat} & interactive demo \\
HobbyLM-1B code & \code{github.com/fuellabs-ai/hobbylm-1b} & model code, GGUF converter, runtime patches, computer-use app \\
\bottomrule
\end{tabularx}
\end{table}

\section{Training configuration}\label{app:config}
The pretraining runs (Table~\ref{tab:hp-pre}) use the optimizer and schedule of Section~\ref{sec:arch}.
All post-training runs (Table~\ref{tab:hp-post}) use AdamW with $\beta=(0.9,0.95)$ and no weight decay, a linear warmup, and a cosine decay to zero.
The configuration of HobbyLM-1B is in Table~\ref{tab:config1b}, and that of HobbyLM-Image in Table~\ref{tab:hp-img}.
Where the repository does not record a value used by a specific run, we give the training script's default and mark it with $^\dagger$. Values that are neither recorded nor defaulted are marked ``--''.

\begin{table}[h]
\centering\small
\caption{Pretraining and context-extension configuration (8$\times$H100, DDP, \code{torch.compile}).}\label{tab:hp-pre}
\begin{tabular}{@{}l r r r r@{}}
\toprule
 & 130M & 500M & 500M ctx-2k & 500M ctx-8k \\
\midrule
Initialization & random & random & 500M & 500M ctx-2k \\
Steps & 9,537 & 38,147 & 2,000 & -- \\
Tokens per step & 1,048,576 & 1,048,576 & 1,048,576 & -- \\
Tokens & ${\approx}10$B & ${\approx}40$B & ${\approx}2.1$B & -- \\
Sequence length & 1,024 & 1,024 & 2,048 & 8,192 \\
RoPE $\theta$ & $10^4$ & $10^4$ & $10^4$ & $10^6$ \\
Peak LR (Muon / AdamW) & 0.04 / $6\times10^{-4}$ & 0.04 / $6\times10^{-4}$ & $\times0.5$ & -- \\
Warmup / decay & \multicolumn{4}{c}{100 steps / linear over the last 40\% to $0.1\times$ peak} \\
Muon & \multicolumn{4}{c}{Nesterov 0.95, weight decay 0.1, 5 Newton--Schulz steps (bf16)} \\
AdamW & \multicolumn{4}{c}{$\beta=(0.9,0.95)$, weight decay 0.1} \\
Gradient clipping & \multicolumn{4}{c}{1.0} \\
Expert bias & \multicolumn{4}{c}{$u=10^{-3}$ per forward pass, frozen after 95\% of steps} \\
Precision & \multicolumn{4}{c}{fp32 master weights, bf16 autocast, TF32, fp32 router} \\
Step time & 742\,ms & 1,208\,ms & -- & -- \\
Final validation loss & 3.3016 & 3.0281 & 2.999 & 2.94 \\
\bottomrule
\end{tabular}
\end{table}

\begin{table}[h]
\centering\small
\caption{Post-training configuration. All runs use 8$\times$H100 unless noted. Micro-batch is per GPU.}\label{tab:hp-post}
\begin{tabularx}{\linewidth}{@{}l Y r r r r@{}}
\toprule
Stage & Trainable / data & LR & Warmup & Micro & Steps \\
\midrule
Chat SFT & full model; SmolTalk (${\approx}1.5$M) & $2\times10^{-5\dagger}$ & 150$^\dagger$ & 8$^\dagger$ & 28,000 \\
Chat DPO & full model; UltraFeedback, $\beta=0.1$ & $5\times10^{-7}$ & -- & 2$^\dagger$ & 2,000$^\dagger$ \\
Diffusion chat SFT & full model; SmolTalk, sequence 1,024 & $2\times10^{-5}$ & 150$^\dagger$ & 16 & 8,000 \\
Vision stage 1 & projector; LCS-558K & $10^{-3\dagger}$ & 100$^\dagger$ & 32$^\dagger$ & 1,500 \\
Vision stage 2 & model + projector; LLaVA-Instruct-150K & $2{\times}10^{-5}$ / $10^{-4}$ & 60$^\dagger$ & 8 & 2,500 \\
Sound stage 1 (1$\times$H100) & audio projector; Clotho & -- & -- & -- & 1,200 \\
Speech stage 1 (1$\times$H100) & speech projector; LibriSpeech-100 & -- & -- & -- & 1,500 \\
Spoken QA (1$\times$H100) & model + speech projector; VoiceAssistant-400K & -- & -- & -- & 1,200 \\
joint5 & model + all projectors; 5 paths & $2{\times}10^{-5}$ / $10^{-4}$ & 80$^\dagger$ & 4 & 2,000 \\
joint6 / joint7 & + tools / + agent loops & $2{\times}10^{-5}$ / $10^{-4}$ & 80$^\dagger$ & 4$^\dagger$ & 2,400 / 3,200 \\
Tool SFT & full model; Nemotron (268K) & $2\times10^{-5}$ & 150$^\dagger$ & 8 & 4,000 \\
Tool SFT, combined & full model; Nemotron + BitAgent + multi-call (${\approx}816$K) & $2\times10^{-5\dagger}$ & 150$^\dagger$ & 8$^\dagger$ & 6,000 \\
\bottomrule
\end{tabularx}
\end{table}

\begin{table}[h]
\centering\small
\caption{HobbyLM-Image configuration.}\label{tab:hp-img}
\begin{tabularx}{\linewidth}{@{}l Y@{}}
\toprule
Backbone & DiT, 16 layers, width 1024, 16 heads, MLP ratio 3, patch 1, 333M parameters \\
Input & two-panel DC-AE f32c32 latent plus 2 mask channels (34 channels) \\
Conditioning & CLIP ViT-L/14 text features (64 tokens $\times$ 768), timestep, task \\
Objective & rectified flow, logit-normal $t$, foreground weight $1+1.5m$ for edits \\
Optimizer & AdamW, $\beta=(0.9,0.95)$, weight decay 0.02, gradient clipping 1.0 \\
Learning rate & $5\times10^{-5}$ from scratch, $2\times10^{-5}$ continued; warmup 1,000 / 500 steps; cosine to 10\% \\
Micro-batch & 32 (512px pilot), 64 (512px continued), 16 (1024px) \\
Regularization & EMA 0.999; per-sample loss clamp 4.0; text dropout 10\%; mask dropout 50\% \\
Sampling & Euler, about 100 steps, CFG about 5 \\
\bottomrule
\end{tabularx}
\end{table}

\section{Compute}\label{app:compute}
Table~\ref{tab:compute} lists the compute recorded for the main runs.
All 500M-family and image work ran on Modal serverless GPUs.
Short runs (projector alignment, ablations, evaluations) are not listed individually.

\begin{table}[h]
\centering\small
\caption{Recorded compute for the main runs. GPU-hours are wall-clock time multiplied by the number of GPUs; where only the cost was recorded, we give the cost.}\label{tab:compute}
\begin{tabular}{@{}l l r r@{}}
\toprule
Run & Hardware & Wall-clock & GPU-hours / cost \\
\midrule
130M pretraining (10B tokens) & 8$\times$H100 & ${\approx}2$\,h & ${\approx}16$ \\
500M pretraining (40B tokens) & 8$\times$H100 & ${\approx}13$\,h & ${\approx}104$ \\
500M context extension to 2k & 8$\times$H100 & ${\approx}1$\,h & ${\approx}8$ \\
Tool SFT (4,000 steps) & 8$\times$H100 & ${\approx}23$\,min & ${\approx}3$ \\
joint co-training (1,600 steps) & 8$\times$H100 & ${\approx}40$\,min & ${\approx}5$ \\
Diffusion LM adaptation (${\approx}20$B tokens) & 8$\times$B200 & -- & ${\approx}$\$155 \\
HobbyLM-Image (all experiments) & H100 / B200 & -- & ${\approx}$\$445 \\
\midrule
HobbyLM-1B (100B tokens) & 4$\times$H200 & ${\approx}76$\,h & ${\approx}304$ \\
\bottomrule
\end{tabular}
\end{table}

\section{Evaluation details}\label{app:eval}

\paragraph{Language modeling.}
For the 130M and 500M models and the reference models (Table~\ref{tab:lmeval}), we use \code{lm-evaluation-harness} v0.4.9.1~\citep{gao2024lmeval}, 0-shot, through a custom \code{LM} wrapper.
The wrapper scores log-likelihoods with the MoE and the GPT-2 tokenizer. It encodes context and continuation jointly, matching the harness's handling of Hugging Face models; this fix changed averages by at most 0.08 points.
We report \code{acc\_norm} where the task defines it and \code{acc} otherwise.
Reference models were loaded from Hugging Face and run through the same task configuration.
HobbyLM-1B and HobbyLM-1B-Instruct were evaluated by their builders with a separately recorded setup: lm-eval 0.4.13, fp32, 0-shot, all examples of each split, with the same \code{acc\_norm}/\code{acc} choices. The wrapper description above does not apply to them. No complete training-data exposure audit was done for these lm-eval tasks.

\paragraph{Vision--language.}
We run \code{lmms-eval}~\citep{zhang2024lmmseval} with a custom model adapter.
The image sentinel is placed before the prompt \code{USER: <question>\textbackslash nASSISTANT:}.
Decoding is greedy with a repetition penalty of 1.3 and no repeated trigrams, with at most 8 new tokens for POPE and 32 otherwise.
\begin{itemize}
\item \emph{Normalized exact match} takes the leading yes/no for POPE. For GQA and VQAv2 it takes the salient content word that does not occur in the question.
\item \emph{Containment} generates the full answer and counts it correct when the normalized gold answer appears in it.
\end{itemize}

\paragraph{Function calling.}
The needle-style metrics~\citep{cactus_needle} compare a predicted JSON call list with the reference:
\begin{itemize}
\item JSON parse rate;
\item name F1 over tool names;
\item argument accuracy;
\item value accuracy;
\item parameter hallucination: the rate of argument keys absent from the schema;
\item exact match of the full call list.
\end{itemize}
BFCL v3~\citep{patil2025bfcl} uses AST matching. Every predicted parameter must exist in the reference, and each value must be in the reference's list of acceptable values. Parallel calls are matched without regard to order. \emph{Relevance} requires a call, and \emph{irrelevance} requires abstention.
Computer-use splits are scored with the same metrics on held-out rows, split by screen (single step) or by trajectory (planning).

\section{HobbyLM-1B-Instruct training details}\label{app:sft1b}
This appendix was prepared from the 1B authors' investigation of saved checkpoints and configuration files. Counter values are as recorded in the checkpoints, not an independent recount of the remote training stream.

\begin{table}[!htbp]
\centering\small
\caption{Selected checkpoint path of HobbyLM-1B-Instruct. Identity is the SHA-256 prefix of each stage's output. The 800 extension and pilot steps precede the SFT step counter. The step-2,751--4,750 continuation and a separate step-3,400 summarization branch are not ancestors of step 3,450.}\label{tab:lineage1b}
\begin{tabular}{@{}l l r@{}}
\toprule
Stage & Identity & Steps \\
\midrule
Base (step 95,367) & \code{947e80d9} & pretraining \\
2K extension & \code{0add7bf0} & 550 \\
Function-calling pilot & \code{3f60c420} & 100 \\
4K plain-RoPE extension (R2) & \code{4a0c8215} & 150 \\
SFT D: recovery, step 500 & \code{7f1e7213} & 500 \\
SFT E: continuation, step 750 & \code{df65ea5a} & 250 \\
SFT F: continuation, step 2,750 & \code{62bba97b} & 2,000 \\
SFT G: Broad-Instruct release, step 3,450 & \code{ef62430d} & 700 \\
\bottomrule
\end{tabular}
\end{table}

\begin{table}[!htbp]
\centering\small
\caption{Supervised-token increments (SFT plus replay) recorded in the checkpoints. The final raw-position counter is $3{,}617{,}587{,}200 = 3{,}450\times1{,}048{,}576$, which includes padded and masked positions.}\label{tab:sfttokens1b}
\begin{tabular}{@{}l r@{}}
\toprule
SFT stage & Supervised tokens \\
\midrule
D, step 0 $\rightarrow$ 500 & 11,869,621 \\
E, step 500 $\rightarrow$ 750 & 5,789,137 \\
F, step 750 $\rightarrow$ 2,750 & 43,678,658 \\
G, step 2,750 $\rightarrow$ 3,450 & 16,776,587 \\
\midrule
Cumulative & 78,114,003 \\
\bottomrule
\end{tabular}
\end{table}

\begin{table}[!htbp]
\centering\footnotesize
\caption{Prepared datasets and configured sampling shares for the final Broad-Instruct stage (step 2,750 $\rightarrow$ 3,450). Shares are conditional on selecting an SFT row and are rounded; replay is a separate share of about 0.46\% of all sampled rows. Row counts are prepared rows, not unique examples consumed. Earlier stages used different mixtures and, for some buckets, different files. Revision prefixes are as recorded.}\label{tab:sftmix1b}
\begin{tabularx}{\linewidth}{@{}>{\raggedright\arraybackslash}p{3.3cm} Y r r@{}}
\toprule
SFT bucket & Source & Rows & Share \\
\midrule
Nemotron Policy-B & \code{nvidia/Nemotron-SFT-Agentic-v2}; consumed file's revision unknown & 130,411 & 22.6\% \\
Verified math v2 & internal generation & 159,852 & 11.6\% \\
Math answer-key v1 & internal generation & 120,000 & 8.7\% \\
smol-magpie-ultra & SmolTalk, revision \code{5feaf2fd} & 261,204 & 8.5\% \\
Tool trajectory v1 & internal generation & 30,000 & 7.5\% \\
Python coding v2 & internal generation & 89,998 & 6.0\% \\
Multispan synthesis v2 & internal generation & 89,986 & 5.4\% \\
Semantic abstention & upstream sources not fully traced & 5,455 & 5.2\% \\
Document knowledge v1 & internal generation & 77,727 & 4.7\% \\
Python coding v1 & internal generation & 60,000 & 4.0\% \\
Precise instruction v2 & internal generation & 48,236 & 3.6\% \\
Callable positives & Glaive function-calling v2 / ToolACE / xLAM-60k, revisions \code{e7f4b645} / \code{6bda777c} / \code{26d14ebf} & 17,994 & 3.1\% \\
smol-rewrite & SmolTalk, \code{5feaf2fd} & 4,599 & 2.7\% \\
Structured output v2 & internal generation & 30,702 & 2.3\% \\
Grounded summarization & SmolTalk smol-summarize, \code{5feaf2fd} & 2,143 & 1.5\% \\
Short-context tool result & internal generation & 1,978 & 1.3\% \\
Tool-result natural answer & internal generation & 654 & 0.5\% \\
Grounded JSON extraction & smol-summarize emails, constructed targets & 528 & 0.45\% \\
smol-constraints & SmolTalk, \code{5feaf2fd} & 5,400 & 0.4\% \\
\bottomrule
\end{tabularx}

\smallskip
\parbox{\linewidth}{\footnotesize The final stage rebuilt the magpie, rewriting, summarization and extraction buckets; earlier magpie, rewriting and extraction counts were 261,254, 4,624 and 548. Internal generators have recorded seed tags, but their code versions are not linked. Replay uses pretraining-style shards whose file list and revisions were not audited. A configured share is neither a measured supervised-token share nor a measure of a bucket's contribution to learning.}
\end{table}

\paragraph{Batch and replay.}
Each optimizer step uses 256 sequences of 4,096 tokens (micro-batch 8, gradient accumulation 32). Pretraining-style replay is configured at about 0.46\% of sampled rows, with an expected contribution of about 20\% of supervised tokens; its actual supervised-token share was not measured separately.

\paragraph{Objective.}
In the inspected training implementation, SFT supervision covers the response, including its end-of-text token, and replay rows are supervised on all tokens. With replay weight $\lambda=1$, the chunked cross-entropy is normalized over the combined supervised tokens, and the auxiliary and z-loss terms are added separately. The displayed logging formula differs from this backward objective, so we describe the inspected implementation.

\paragraph{Optimizer.}
Muon updates the matrix parameters and AdamW the rest, with nominal peak learning rates of $8\times10^{-4}$ and $1.2\times10^{-5}$, AdamW $\beta=(0.9,0.95)$ and weight decay 0.1.
Stages D and E end at about $0.1\times$ the nominal rates, and F and G at $0.5\times$; the final rates are $4\times10^{-4}$ (Muon) and $6\times10^{-6}$ (AdamW).
Equal rates at G's endpoints do not show that the rate was constant during G; G's full schedule and launch flags are unknown.

\paragraph{Execution caveats.}
The remote trainer's runtime hash was not recorded. Code history supports the reconstruction, but does not establish byte-for-byte identity with what ran. Part of the sampler coverage comes from a reconstructed step-500 data state, and some buckets have unknown wrap counts, so per-bucket coverage is not exact measured consumption.

\section{Prompt formats}\label{app:prompts}
The formats below are those of the 130M/500M family (Sections~\ref{sec:500m}--\ref{sec:mm}); HobbyLM-1B-Instruct uses different formats, given at the end of this appendix. All use plain-text role prefixes rather than special tokens. In the 500M training data, each assistant turn ends with the end-of-text token.

\paragraph{Chat.}
\begin{lstlisting}
SYSTEM: <optional system prompt>
USER: <message>
ASSISTANT: <response><|endoftext|>
\end{lstlisting}

\paragraph{Function calling.}
\begin{lstlisting}
TOOLS: [{"name": ..., "description": ..., "parameters": {...}}, ...]
USER: <request>
ASSISTANT: [{"name": "<tool>", "arguments": {...}}]<|endoftext|>
TOOL: <result>
ASSISTANT: <answer><|endoftext|>
\end{lstlisting}

\paragraph{Computer use (planning).}
\begin{lstlisting}
SCREEN:
[Window] "Calculator"
[Button] "Seven"
...
GOAL: <goal>
DONE: <instruction 1>; <instruction 2>   (or: nothing yet)
NEXT: [{"name": "click", "arguments": {"name": "Seven", "control_type": "Button"}}]
\end{lstlisting}

\paragraph{Multimodal.}
The projected features replace a sentinel token (image 50257, sound 50258, video 50261, speech 50262) placed before the text turn:
\begin{lstlisting}
<IMAGE> USER: <question>
ASSISTANT: <answer><|endoftext|>
\end{lstlisting}

\paragraph{HobbyLM-1B-Instruct.}
Training formats vary across the 1B SFT data buckets. The bundled chat template follows the role-marker convention of the inspected conversational and tool-use examples. Rewriting, summarization and JSON-extraction examples use instruction-and-input formats without these markers, and format verification does not cover every bucket.
The role markers are ordinary tokenizer text, not dedicated special tokens. An inspected tool-result example:
\begin{lstlisting}
<|user|>
Convert 32.0 miles to kilometers.
<|assistant|>
[convert_length(value=32.0, from_unit='miles', to_unit='km')]
<|tool|>
{"value": 32.0, "from_unit": "miles", "to_unit": "km", "converted": 51.5}
<|assistant|>
That's roughly 51.5 km.<|endoftext|>
\end{lstlisting}
In this example only the final natural-language answer and the end-of-text token are supervised; the earlier assistant and tool text is context. This does not mean every assistant turn is unsupervised in every bucket.

\section{Author contributions}\label{app:contrib}
\textbf{Harish Santhanalakshmi Ganesan} designed the HobbyLM architecture and ran the ablation study. Harish Santhanalakshmi Ganesan also:
\begin{itemize}
\item trained the 130M and 500M models, including context extension;
\item built all 500M post-training: the chat, diffusion-LM, SAE, multimodal omni, function-calling and computer-use models;
\item wrote the GGUF export, the \code{hobby-rs} engine, the \code{hobby-chat} application and the demo Space;
\item designed and trained HobbyLM-Image;
\item wrote Sections~\ref{sec:intro}--\ref{sec:mm} and~\ref{sec:image}.
\end{itemize}
\textbf{Hariharan M.\,S.} and \textbf{Prabhurajhan R.} adopted the architecture to build HobbyLM-1B. They:
\begin{itemize}
\item curated the 100B-token training data;
\item built the data pipeline and training infrastructure;
\item ran the pretraining and anneal phases;
\item built HobbyLM-1B-Instruct and its GGUF release;
\item carried out the evaluation and analysis reported in Section~\ref{sec:1b}, which is based on their draft.
\end{itemize}
Prabhurajhan R. developed the HobbyLM-1B MLX implementation for Apple Silicon.
Prabhurajhan R. built HobbyLM-1B Computer Use v0.1.

\end{document}